\documentclass[twoside]{article}
\usepackage[accepted]{aistats2026}

\usepackage[round]{natbib}

\usepackage[utf8]{inputenc} 
\usepackage[T1]{fontenc}    
\usepackage{hyperref}       
\usepackage{url}            
\usepackage{booktabs}       
\usepackage{amsfonts}       
\usepackage{nicefrac}       
\usepackage{microtype}      
\usepackage{xcolor}         

\usepackage{hyperref}
\usepackage{graphicx}
\usepackage{algorithmic}
\usepackage{algorithm}

\usepackage{amsmath}
\usepackage{amssymb}
\usepackage{mathtools}
\usepackage{amsthm}
\usepackage{enumitem}
\usepackage{todonotes}

\usepackage{pifont}
\newcommand{\cmark}{\ding{51}}%
\newcommand{\xmark}{\ding{55}}%

\newtheorem{theorem}{Theorem}[section]

\newtheorem{lemma}[theorem]{Lemma}

\theoremstyle{definition}

\theoremstyle{remark}
\newtheorem{remark}[theorem]{Remark}

\usepackage[capitalize,noabbrev]{cleveref}

\hypersetup{%
    colorlinks=true,
    linkcolor=mydarkblue,
    citecolor=mydarkblue,
    filecolor=mydarkblue,
    urlcolor=mydarkblue,
}
\definecolor{mydarkblue}{rgb}{0,0.08,0.45}

\usepackage[frozencache,cachedir=.]{minted}
\definecolor{LightGray}{gray}{0.9}

\hypersetup{%
    colorlinks=true,
    linkcolor=mydarkblue,
    citecolor=mydarkblue,
    filecolor=mydarkblue,
    urlcolor=mydarkblue,
}

\begin{document}
\runningtitle{Q-Learning with Shift-Aware Upper Confidence Bound in Non-Stationary Reinforcement Learning}
\twocolumn[
\aistatstitle{Q-Learning with Shift-Aware Upper Confidence Bound in Non-Stationary Reinforcement Learning}
\vspace{-0.2in}
\aistatsauthor{Ha Manh Bui \And Felix Parker \And Kimia Ghobadi \And Anqi Liu}
\aistatsaddress{Johns Hopkins University, Baltimore, MD, U.S.A. \\
\texttt{\{hbui13, fparker9, kimia, aliu.cs\}@jhu.edu}

}
]

\begin{abstract}
We study the Non-Stationary Reinforcement Learning (RL) under distribution shifts in both finite-horizon episodic and infinite-horizon discounted Markov Decision Processes (MDPs). In the finite-horizon case, the transition functions may suddenly change at a particular episode. In the infinite-horizon setting, such changes can occur at an arbitrary time step during the agent's interaction with the environment. While the Q-learning Upper Confidence Bound algorithm (QUCB) can discover a proper policy during learning, due to the distribution shifts, this policy can exploit sub-optimal rewards after the shift happens. To address this issue, we propose Density-QUCB (DQUCB), a shift-aware Q-learning~UCB algorithm, which uses a transition density function to detect distribution shifts, then leverages its likelihood to enhance the uncertainty estimation quality of Q-learning~UCB, resulting in a balance between exploration and exploitation. Theoretically, we prove that our oracle DQUCB achieves a better regret guarantee than QUCB. Empirically, our DQUCB enjoys the computational efficiency of model-free RL and outperforms QUCB baselines by having a lower regret across RL tasks, as well as a COVID-19 patient hospital allocation task using a Deep-Q-learning architecture.
\end{abstract}
\section{Introduction}
Non-stationary Reinforcement Learning (RL) is a sequential decision-making problem in which an agent interacts with a changing environment over time to maximize rewards~\citep{sutton1998RL}. One instance of this setting is considering a run of Markov Decision Process (MDP) dynamics in the finite-horizon episodic MDP~\citep{jin2018isQ, menard2021UCBmomentum}, where the reward and transition functions can change within an episode. To handle this case, \citet{jin2018isQ} introduces the Q-learning~UCB (i.e., QUCB) algorithm by combining Q-learning~\citep{watkins1992QLearning} with the idea of Upper Confidence Bound (UCB) in bandit literature~\citep{kearns2002nearoptimal,lattimore2017bandit}. On the one hand, Q-learning tries to estimate the optimal Quality(Q)-function, i.e., the state-action value function. With a Q-function, at every state, the agent can greedily select the action with the largest Q-value to interact with the environment. On the other hand, the UCB term (calculated based on the number of visited state-action pairs) measures the uncertainty of the Q-function, aiming to balance the exploration and exploitation performance. Intuitively, when the UCB term is large, the model will be more uncertain. Conversely, when the UCB term is small, the model will be more certain about the Q-values. By leveraging this uncertainty information of UCB, \citet{jin2018isQ,wang2020Qlearning} show that QUCB is provably efficient in non-stationary RL. Although having a higher regret than model-based RL (e.g., UCBVI, UCBMQ, etc.), QUCB is model-free and typically simpler, enjoys better time and space complexity, and thus is more prevalent in infinite-state environments with modern Deep-RL architectures~\citep{mnih2015humanlevel,wang2020Qlearning,chai2025empowering}.

\begin{table}[t!]
    \centering
    \scalebox{0.78}{
    \begin{tabular}{cccc}\\
    \toprule  
    Method & $\begin{matrix}
    \text{Shift-Aware} \\
    \text{ability}
    \end{matrix}$ & $\begin{matrix}
    \text{Computational} \\
    \text{efficiency}
    \end{matrix}$ & $\begin{matrix}
    \text{Handle infinite-state} \\
    \text{\& Deep-RL}
    \end{matrix}$\\\midrule
    QUCB & \xmark & \cmark & \cmark\\
    UCBVI  & \cmark & \xmark & \xmark\\
    UCBMQ & \cmark & \xmark & \xmark\\
    Ours & \cmark & \cmark & \cmark\\
    \bottomrule
    \end{tabular}}
    \caption{A comparison between methods regarding shift-aware ability, computational efficiency (time \& space), and ability to handle infinite-state \& Deep-RL.}
    \label{tab:teaser}
\end{table}

That said, QUCB and the methods mentioned above only consider the finite-horizon episodic MDP setting, where reward and transition functions can change within an episode but are fixed across different episodes. Similarly, in the infinite-horizon discounted MDP, \citet{wang2020Qlearning,yang2021qlearning} assume the transition function is fixed across time steps. This is a strong assumption because the transition function can suddenly change in the real world. Therefore, we study a more challenging and realistic scenario by considering non-stationary RL under distribution shifts, i.e., the transition function can additionally change at a particular episode in the finite-horizon episodic MDP, or at a time step in the infinite-horizon discounted MDP. These settings are critical and align with real-world applications. E.g., in robot navigation, over multiple episodes (interaction trials with the environment), a robot needs to learn an optimal route to the final destination, as shown in Fig.~\ref{fig:setting}. There are slippery sections on the route that may cause the robot to move in an undesirable direction. Over time, the environment, i.e., the slippery levels, can change across episodes due to weather conditions. Because of this distribution shift, the optimal policy can change (e.g., route A with more slippery than B, now becomes less slippery than B). As not designed to handle this kind of shift, QUCB with non-shift-aware ability, may exploit sub-optimal reward by an old policy (learned policy before the shift), resulting in a bad performance after the shift occurs.

To address the non-shift-aware issue in QUCB, we introduce Density-QUCB (DQUCB), a shift-aware Q-learning~UCB algorithm, which uses a transition density function to adjust the exploration rate of the UCB term. The key idea of our method is leveraging the likelihood of this density function as an out-of-distribution detector, i.e., when there is no shift, its likelihood will be high, and our model will be more certain (exploitation). In contrast, its likelihood will be low when the distribution shift occurs, which means our model will be more uncertain (exploration). This helps our model improve UCB uncertainty quantification, be able to explore new policies, and avoid exploiting an old policy when the shift happens, resulting in a better regret guarantee than the non-shift-aware QUCB algorithm.

Our contributions are outlined in Tab.~\ref{tab:teaser} and as follows:
\begin{itemize}[leftmargin=*]
    \item We propose DQUCB, a shift-aware Q-learning~UCB algorithm to enhance the uncertainty quantification quality of UCB with Q-learning, resulting in a balance between exploration and exploitation in the non-stationary RL under distribution shifts.
    \item We theoretically prove the regret of our DQUCB algorithm in the finite-horizon episodic MDP is at most $\mathcal{O}\left(\sqrt{H^5 (1+\epsilon)^2 |\mathcal{S}| |\mathcal{A}| K \log(|\mathcal{S}| |\mathcal{A}|KH/\delta)}\right)$, where $|\mathcal{S}|$, $|\mathcal{A}|$, $H$, $K$, $\epsilon$ are the number of state, action, planning horizon, episodes, and density estimator error, correspondingly. And in infinite-horizon discounted MDP is at most $\mathcal{O}\left(\frac{|\mathcal{S}| |\mathcal{A}|(1+\epsilon)}{(1-\gamma)^6\min\{\Delta_{\min}, \bar{\Delta}_{\min}\}} \log\left(\frac{|\mathcal{S}| |\mathcal{A}|T}{(1-\gamma)\min\{\Delta_{\min}, \bar{\Delta}_{\min}\}}\right)\right)$, where $\gamma$ is the discount factor, $T$ is the number of cumulative steps, and $\Delta_{\min}, \bar{\Delta}_{\min}$ are the minimum sub-optimality gap. Notably, our oracle DQUCB algorithm (i.e., when $\epsilon \rightarrow 0$) is strictly better than the non-shift-aware version (i.e., QUCB) in both finite-horizon episodic and infinite-horizon discounted MDP settings.
    \item We empirically show DQUCB outperforms QUCB by achieving lower regret on GridWorld and Frozen-Lake. In addition, it exhibits the computational efficiency of model-free RL and outperforms model-based RL (e.g., UCBVI, UCBMQ) in space and time complexity (see Fig.~\ref{fig:complexity}). Furthermore, we extend this framework to neural-network versions on CartPole, COVID-19 patient hospital allocation task, and show it consistently achieves a lower cumulative regret than other Deep Q-learning baselines.
\end{itemize}

\section{Preliminary}
In this paper, we consider two RL settings in an unknown MDP, including when the learning agent interacts with the system with (i.e., finite-horizon episodic MDP) and without (i.e., infinite-horizon discounted MDP) any reset. For simplicity, our formal setting considers one time shift. Our results can also be extended to multiple time shifts using our general proof techniques or a similar analysis. We formally define these settings under distribution shifts as follows:

\subsection{Finite-horizon episodic MDP under distribution shifts}
In the first setting, an episodic Markov Decision Process (MDP) is a tuple $\mathcal{M}:=\langle \mathcal{S}, \mathcal{A}, H, \mathbb{P}, r \rangle$, where $\mathcal{S}$ is the state space, $\mathcal{A}$ is the action space, $H\in \mathbb{N}$ is the planning horizon, $\mathbb{P}_h: \mathcal{S}\times \mathcal{A} \rightarrow \Delta(\mathcal{S})$ is the transition operator at step $h$ that takes a state-action pair and returns a distribution over states, and $r_h: \mathcal{S} \times \mathcal{A} \rightarrow [0,1]$ is the
deterministic reward function at step $h$. For a finite-horizon episodic MDP, the agent interacts with the MDP for $K \in \mathbb{N}$ episodes. For each episode $k\in [K]$, the agent starts at an initial state $s_1 \in \mathcal{S}$ picked by an adversary~\citep{jin2018isQ,menard2021UCBmomentum}. In this setting, we focus on the distribution shift of the transition operator $\mathbb{P}_h$ at a particular episode $\bar{K} \in [K]$. Let us denote $\bar{\mathbb{P}}_h$ as the transition operator after the distribution shift occurs. Hence, $\bar{s}_h$ means the state r.v. sampled from $\bar{\mathbb{P}}_h$. Given a policy $\pi$ which is a sequence of mappings $\pi_h:\mathcal{S} \rightarrow \mathcal{A}$ for $h\in [H]$, for a state $s\in \mathcal{S}$, the value function of state $s\in \mathcal{S}$ at the $h$-step are defined as
\begin{align}
    V_h^\pi(s) := \mathbb{E}\left[\sum_{h'=h}^H r_{h'}(s_{h'}, \pi_{h'}(s_{h'})) \Big| s_h=s\right]\nonumber,\\
    \quad \bar{V}_h^\pi(s) := \mathbb{E}\left[\sum_{h'=h}^H r_{h'}(\bar{s}_{h'}, \pi_{h'}(\bar{s}_{h'})) \Big| \bar{s}_h=s\right],
\end{align}
and the associated $Q$-function of a state-action pair $(s,a)\in \mathcal{S}\times \mathcal{A}$ at the $h$-step are $Q_h^\pi(s,a) := r_h(s,a) + \mathbb{E}\left[\sum_{h'=h+1}^H r_{h'}(s_{h'}, \pi_{h'}(s_{h'})) \Big| s_h=s, a_h=a\right]$ and $\bar{Q}_h^\pi(s,a) := r_h(s,a) + \mathbb{E}\left[\sum_{h'=h+1}^H r_{h'}(\bar{s}_{h'}, \pi_{h'}(\bar{s}_{h'})) \Big| \bar{s}_h=s, a_h=a\right]$.
We let $\pi^*$ and $\bar{\pi}^*$ be the optimal policy s.t. $V^{\pi^*}(s)=V^*(s)=\sup_\pi V^\pi(s)$ and $\bar{V}^{\bar{\pi}^*}(s)=\bar{V}^*(s)=\sup_\pi \bar{V}^\pi(s)$. Hence, $Q^*(s,a)$ and $\bar{Q}^*(s,a)$ mean the Q-function under optimal policy $\pi^*$ and $\bar{\pi}^*$. respectively, $\forall (s,a)$. For each episode $k\in [K]$, the learning agent specifies a policy $\pi^k$, plays $\pi^k$ for $H$ steps and observes trajectory $(s_1,a_1), \cdots, (s_H,a_H)$. The total number of steps is $KH$, and the total (expected) regret of an execution instance of the agent is then
\small
\begin{align}
    \text{Regret}(K) &:= \sum_{k=1}^{\bar{K}} (V_1^* - V_1^{\pi_k})(s_1^k) + \sum_{k=\bar{K}+1}^{K} (\bar{V}_1^* - \bar{V}_1^{\pi_k})(s_1^k).
\end{align}
\normalsize

\begin{figure}[t!]
    \centering
    \includegraphics[width=1.0\linewidth]{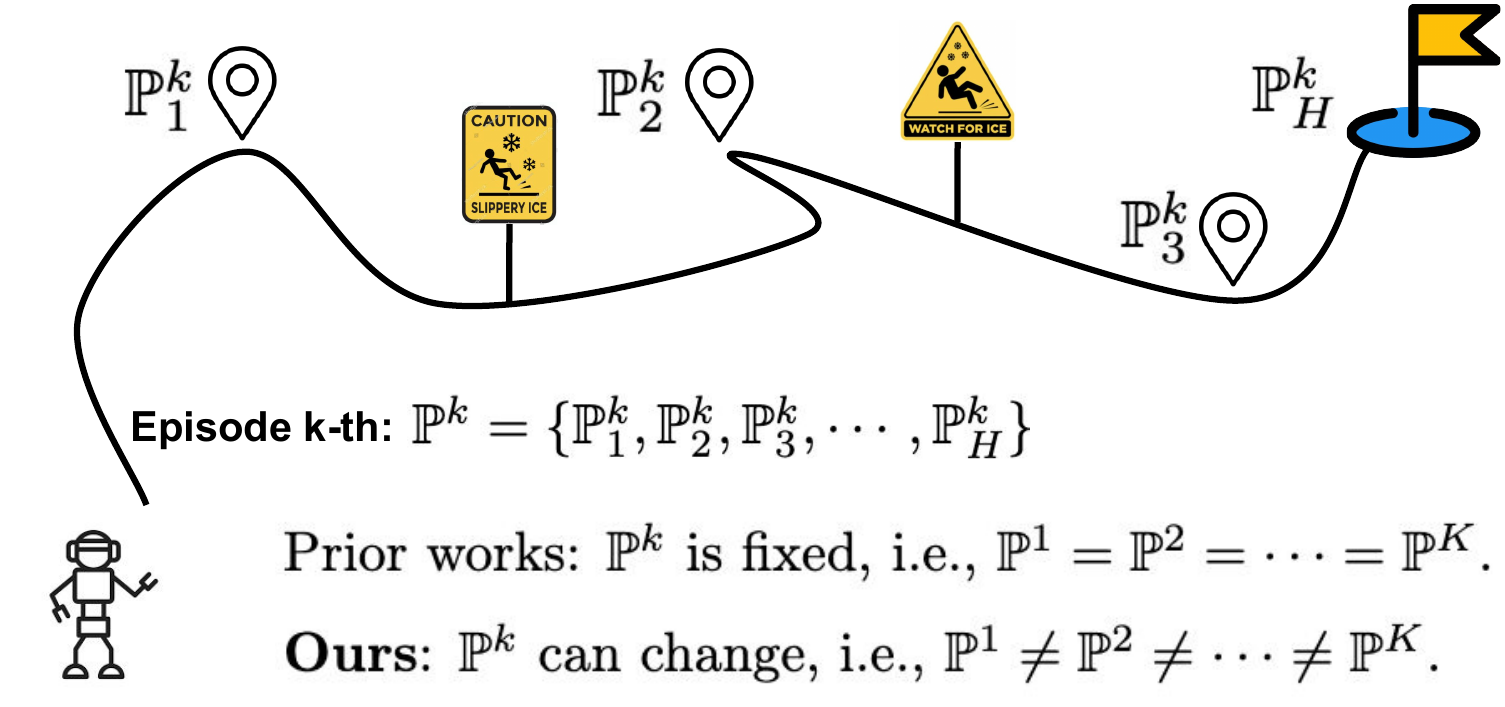}
    \caption{Robot navigation example, where $\mathbb{P}^k$ represents the set of slippery distributions on the route $\{\mathbb{P}_h\}_{h\in [H]}$ at episode k-th. We consider a more general case of prior works. Specifically, prior works set $\{\mathbb{P}_h\}_{h\in [H]}$ stay the same over episode $k \in [K]$, we consider $\{\mathbb{P}_h\}_{h\in [H]}$ can change across different episodes $k \in [K]$. Similarly, in the infinite-horizon discounted MDP, prior works consider a fixed $\mathbb{P}$ across time steps, while we consider that $\mathbb{P}$ can change.}
    \label{fig:setting}
\end{figure}

It is worth noting that we consider a general case, rather than a special case of the literature on a non-stationary MDP~\citep{jin2018isQ, menard2021UCBmomentum}. Specifically, prior works consider the case of transition functions $\mathbb{P}_h$ changing across the time horizon $h\in [H]$ within an episode, i.e., $\mathbb{P}_h$ is different over $h\in [H]$ but the set $\{\mathbb{P}_h\}_{h\in [H]}$ stay the same over episode $k \in [K]$. Meanwhile, our method considers that the transition functions additionally change across episodes, i.e., $\mathbb{P}_h$ differs over $h\in [H]$ and the set $\{\mathbb{P}_h\}_{h\in [H]}$ can also change across different episodes $k \in [K]$ (see Fig.~\ref{fig:setting}).

\subsection{Infinite-horizon discounted MDP under distribution shifts}
In the second setting, we consider a tuple $\mathcal{M}_\gamma:=\langle \mathcal{S}, \mathcal{A}, \gamma, \mathbb{P}, r \rangle$, where $\gamma$ is the discount factor, $\mathbb{P}$ is the  transition operator, and $r$ is the reward function~\citep{wang2020Qlearning,yang2021qlearning}. We focus on the distribution shift of the transition operator $\mathbb{P}$ at a particular time step $\bar{T} \in [T]$, where $T \in \mathbb{N}$ is the number of first time steps. Let us denote $\bar{\mathbb{P}}$ as the transition operator after the distribution shift occurs. Hence, $\bar{s}$ means the state r.v. sampled from $\bar{\mathbb{P}}$. Let $\mathcal{C}:=\{\mathcal{S}\times \mathcal{A}\times [0,1]\}^* \times \mathcal{S}$ be the set of all possible trajectories of any length. A non-stationary deterministic policy $\pi: \mathcal{C}\rightarrow \mathcal{A}$ is a mapping from paths to actions. The $V$ function and $Q$ function are defined as follows
\begin{align}
    V^\pi(s) := \mathbb{E}\left[\sum_{i=1}^\infty \gamma^{i-1} r(s_i, \pi(c_i)) \Big| s_1 =s\right]\nonumber,\\
    \quad \bar{V}^\pi(s) := \mathbb{E}\left[\sum_{i=1}^\infty \gamma^{i-1} r(\bar{s}_i, \pi(\bar{c}_i)) \Big| \bar{s}_1 =s\right],
\end{align}
and $Q^\pi(s,a) := r(s,a) + \mathbb{E}\left[\sum_{i=2}^\infty \gamma^{i-1} r(s_i, \pi(c_i)) \Big| s_1=s,a_1=a\right]$, $\bar{Q}^\pi(s,a) := r(s,a) + \mathbb{E}\left[\sum_{i=2}^\infty \gamma^{i-1} r(\bar{s}_i, \pi(\bar{c}_i)) \Big| \bar{s}_1=s,a_1=a\right]$, where $(c_i:=(s_1, \pi(s_1), r(s_1, \pi(s_1)), \cdots, s_i))$ and $(\bar{c}_i:=(\bar{s}_1, \pi(\bar{s}_1), r(\bar{s}_1, \pi(\bar{s}_1)), \cdots, \bar{s}_i))$. Consider the interaction with the environment that starts at state $s_1$, a learning agent specifies an initial non-stationary policy $\pi_1$. At each time step $t$, the agent takes action $\pi_t(s_t)$, observe $r_t$ and $s_{t+1}$, and updates $\pi_t$ to $\pi_{t+1}$. The total regret of the agent for the first $T$ steps is thus defined as
\small
\begin{align}
    \text{Regret}(T) &:= \sum_{t=1}^{\bar{T}} (V^* - V^{\pi_t})(s_t) + \sum_{t=\bar{T}+1}^T (\bar{V}^* - \bar{V}^{\pi_t})(s_t).
\end{align}
\normalsize
It is worth noting that while \citet{wang2020Qlearning} considers a fixed transition function $\mathbb{P}$ across time steps, we consider a more general case by $\mathbb{P}$ can change at a particular time step $t\in [T]$.
\section{\normalsize{Shift-Aware} \large{Density Q-Learning~UCB}}
To handle the aforementioned settings, based on Q-learning, we next introduce our main methodology.

\begin{figure*}[ht!]
    \centering
    \includegraphics[width=0.8\linewidth]{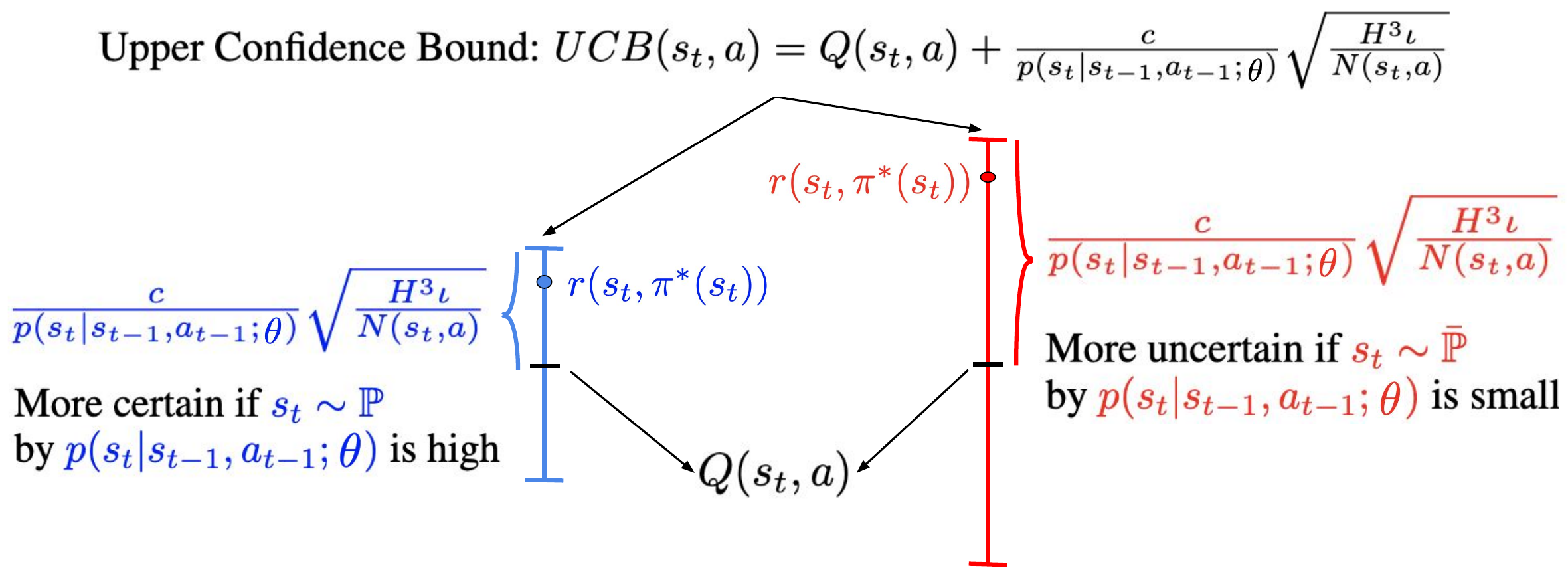}
    \caption{Our shift-aware Q-learning~UCB can be more uncertain (i.e., more exploration) if the environment changes because of a low likelihood from $p(\cdot|s,a;\theta)$, and more certain (i.e., more exploitation) if the environment stays the same because of a high likelihood from $p(\cdot|s,a;\theta)$.}
    \label{fig:framework}
\end{figure*}

\textbf{In the finite-horizon episodic MDP}, our Alg.~\ref {alg:episodic} maintains $Q$ values, $Q_h(s,a)$, for all $(s,a,h)\in \mathcal{S}\times \mathcal{A}\times [H]$ and the corresponding $V$ value, i.e.,
\begin{align}
    V_h(s) \leftarrow \min\{H, \max_{a' \in \mathcal{A}} Q_h(s,a')\}.
\end{align}
If at time step $h\in [H]$, the state is $s\in \mathcal{S}$ and the algorithm takes action $a\in \mathcal{A}$ that maximizes the current estimate $Q_h(s,a)$, and is apprised of the next state $s' \in \mathcal{S}$. The algorithm then updates the $Q$ values by
\begin{align}
    Q_h(s,a) &\leftarrow (1-\alpha_t)Q_h(s,a)\nonumber\\ 
    &\quad + \alpha_t[r_h(s,a)+V_{h+1}(s')+b_t],
\end{align}
where $t=N_h(s,a)$ is the number of times the algorithm has visited the state-action pair $(s,a)$ at $h$, $\alpha_t$ is the learning rate, and $b_t$ is the confidence value indicating how certain the algorithm is about the current state-action pair, and are defined as follows
\begin{align}\label{eq:UCB}
    \alpha_t := \frac{H+1}{H+t}, \quad b_t := \frac{c}{p(s_{h+1}|s_{h}, a_{h};\theta_h)}\sqrt{H^3\iota/t},
\end{align}
where $\iota = \log(|\mathcal{S}| |\mathcal{A}|KH/\delta)$ and $p(s_{h+1}|s_{h}, a_{h};\theta_h)$ is the likelihood value of density function $p(\cdot|s,a;\theta_h)$ parameterized by parameter $\theta_h$, at state $s_{h+1}$ condition on previous state-action $(s_h, a_h)$, and is computed following Bayes's rule as follows
\begin{align}
    p(s_{h+1}|s_{h}, a_{h};\theta_h)=\frac{p(s_{h+1},s_{h}, a_{h};\theta_h)}{p(s_{h}, a_{h};\theta_h)},
\end{align}
where parameter $\theta_h$ is updated by using maximum likelihood estimation with $n$ most recent tuples $(s',s,a)$ from the interaction with the environment. Specifically about our density function $p(s'|s,a;\theta_h)$ above, we use $p(s'|s, a;\theta_h)$ to estimate the likelihood of sample $(s',s,a)$ from the probability density ratio, i.e., $p(s'|s, a;\theta_h) \approx f_X(s'|s,a)/f_Y(s'|s,a)$, where $(s',s,a)\sim \mathbb{P}_Y$ and $\theta_h = \arg\max_{\theta_h}p(D_n|\theta_h)$, with $D_n\sim \mathbb{P}_X$ is the cumulative dataset with $n$ data points. When before the shift, $(s',s,a)\sim \mathbb{P}_h$ and $D_n\sim \mathbb{P}_h$, thus $\mathbb{P}_X =\mathbb{P}_h$ and $\mathbb{P}_Y=\mathbb{P}_h$. When the shift occurs at episode $\bar{K}$, $(s',s,a)\sim \bar{\mathbb{P}}_h$ and the cumulative dataset $D_n\sim \mathbb{P}_h$, thus $\mathbb{P}_X=\mathbb{P}_h$ and $\mathbb{P}_Y=\bar{\mathbb{P}}_h$. After episode $\bar{K}$, the cumulative dataset accumulates data from the new distribution, i.e., $D_n\sim \bar{\mathbb{P}}_h$, hence for $(s',s,a)\sim \bar{\mathbb{P}}_h$, we have $\mathbb{P}_X=\bar{\mathbb{P}}_h$ and $\mathbb{P}_Y=\bar{\mathbb{P}}_h$.

From the uncertainty term $b_t$ in Eq.~\ref{eq:UCB}, we can see that while $t$ is monotonically increasing, $p(s_{h+1}|s_{h}, a_{h};\theta_h)$ is initialized to estimate the density of $(s',s,a)$ sampled from transition operator $\mathbb{P}_h$, over time, the likelihood $p(s_{h+1}|s_{h}, a_{h};\theta_h)$ will be high, yielding the exploration rate in $b_t$ to be low, i.e., the algorithm will be more certain (i.e., exploitation) about its prediction from the mean $Q_h(s,a)$. When the shift happens at episode $\Bar{K}$, $p(s_{h+1}|s_{h}, a_{h};\theta_h)$ behaves as an out-of-distriubiton detector because its likelihood will measure the ratio of $\mathbb{P}_h/\bar{\mathbb{P}}_h$, decreasing value, yielding the exploration rate in $b_t$ will be high, i.e., the algorithm will be more uncertain (i.e., exploration) about its prediction from the mean $Q_h(s,a)$. We show this key idea in Fig.~\ref{fig:framework}.

\begin{remark}\label{rmk:complexity}
\textbf{Comparison with model-based RL}. It is worth noticing that although our function $p(\cdot|s, a;\theta_h)$ estimates the density of samples from the transition function, it does not explicitly model the transition operator $\mathbb{P}_h$ like the model-based method (e.g., UCBVI, UCBMQ, etc.), which needs to store and iterate through all possible $(s',s,a) \in \mathcal{S} \times \mathcal{S} 
\times \mathcal{A}$ tuples with $\Tilde{\mathcal{O}}(KH\mathcal|S|^2|\mathcal{A}|)$ time complexity and $\mathcal{O}(|\mathcal{S}|^2|\mathcal{A}|H)$ space complexity~\citep{auerUCLR2}. Indeed, our training step for $\theta$ of $p(\cdot|s, a;\theta_h)$ only depends on the number of cumulative observed $(s',s,a)$ tuples with $n$ window-size. In general, $p(\cdot|s, a;\theta_h)$ can be any density function. In our experiment, we use Kernel Density Estimation (KDE). Therefore, the time complexity of our DQUCB algorithm is $\mathcal{O}(KHn^2)$, and $\mathcal{O}(|\mathcal{S}||\mathcal{A}|Hn)$, which is much computationally efficient than model-based baselines (e.g., Fig.~\ref{fig:complexity}). Note that we can also further improve our algorithm's complexity by applying other density estimation techniques that have near-linear time and space complexity~\citep{chan2014near,bousquet19optimal,backurs2019space}.
\end{remark}

\begin{figure}[t!]
\begin{algorithm}[H]
    \centering
    \caption{Our DQUCB algorithm in the finite-horizon episodic MDP}\label{alg:episodic}
    \begin{algorithmic}[1]
        \footnotesize
        \STATE \textbf{Initialize} $Q_h(s,a) \leftarrow H$ \& $N_h(s,a)\leftarrow 0$ $\forall (s,a,h)\in \mathcal{S} \times \mathcal{A} \times [H]$, $p(\cdot;\theta_h)$ $\forall h\in [H]$,  $\alpha_t=\frac{H+1}{H+t}$, $\iota = \log(|\mathcal{S}| |\mathcal{A}|KH/\delta)$
        \FOR{every episode $k\in [K]$}
            \STATE Receive $s_0$
            \FOR{step $h\in [H]$}
                \STATE Take $a_h\leftarrow \arg\max_{a'\in \mathcal{A}}Q_h(s_h,a')$, observe $s_{h+1}$
                \STATE $t=N_h(s_h,a_h) \leftarrow N_h(s_h,a_h)+1$
                \STATE $b_t \leftarrow \frac{c}{p(s_{h+1}|s_{h}, a_{h};\theta_h)}\sqrt{H^3\iota/t}$
                \STATE $Q_h(s_h,a_h) \leftarrow (1-\alpha_t) Q_h(s_h,a_h) + \alpha_t\left[r_h(s_h,a_h) + V_{h+1}(s_{h+1}) + b_t\right]$
                \STATE $V_h(s_h) \leftarrow \min\{H,\max_{a'\in \mathcal{A}} Q_h(s_h,a')\}$
                \STATE Update $\theta_h$ of model $p(\cdot;\theta_h)$ by $p(s_{h+1}|s_h, a_h;\theta_h)$
            \ENDFOR
        \ENDFOR
    \end{algorithmic}
\end{algorithm}
\end{figure}

\textbf{In the infinite-horizon discounted MDP}, our Alg.~\ref {alg:infinite} maintains $Q$ values, $Q(s,a)$, for all $(s,a)\in \mathcal{S}\times \mathcal{A}$ and the corresponding $\hat{V}$ value, i.e.,
\begin{align}
    &\hat{V}(s_{t+1}) \leftarrow \max_{a' \in \mathcal{A}} \hat{Q}(s_{t+1},a')\nonumber,\\ 
    &\hat{Q}(s_t,a_t) \leftarrow \min\left\{\hat{Q}(s_t,a_t), Q(s_t,a_t)\right\}.
\end{align}
If, at time step $t\in [T]$, the state is $s\in \mathcal{S}$, the algorithm takes the action $a\in \mathcal{A}$ that maximizes the current estimate $Q(s,a)$, and is apprised of the next sate $s' \in \mathcal{S}$. The algorithm then updates the $Q$ values as follows
\begin{align}
    Q(s_t,a_t) &\leftarrow (1-\alpha_k) Q(s_t,a_t)\nonumber\\
    &\quad + \alpha_k\left[r(s_t,a_t) + \gamma \hat{V}(s_{t+1}) + b_k\right],
\end{align}
where $k=N(s,a)$ is the number of times the algorithm has visited the state-action pair $(s,a)$, $\alpha_k$ is the learning rate, and $b_k$ is the confidence value indicating how certain the algorithm is about the current state-action pair, and are defined as follows
\begin{align}\label{eq:UCB_inf}
    \alpha_k := \frac{H+1}{H+k}, b_k := \frac{c_2}{(1-\gamma)p(s_{t+1}|s_{t}, a_{t};\theta)}\sqrt{H \iota(k)/k},
\end{align}
where $H \leftarrow \frac{\ln(2/(1-\gamma))}{\ln(1/\gamma)}$, $\iota(k) = \log(|\mathcal{S}| |\mathcal{A}|T(k+1)(k+2))$, and $p(s_{t+1}|s_{t}, a_{t};\theta)$ is the likelihood value of density function $p(\cdot|s,a;\theta)$ at state $s_{t+1}$ condition on previous state-action $(s_t, a_t)$, and is computed following Bayes's rule as follows
\begin{align}
    p(s_{t+1}|s_{t}, a_{t};\theta)=\frac{p(s_{t+1},s_{t}, a_{t};\theta)}{p(s_{t}, a_{t};\theta)},
\end{align}
where parameter $\theta$ is updated by $n$ most recent tuples $(s',s,a)$. From the uncertainty term $b_k$ in Eq.~\ref{eq:UCB_inf}, we can see that while $k$ is monotonically increasing, $p(s_{t+1}|s_{t}, a_{t};\theta)$ is initialized to estimate the density of $(s',s,a)$ sampled from transition operator $\mathbb{P}$, over time, the likelihood $p(s_{t+1}|s_{t}, a_{t};\theta)$ will be high, yielding the exploration rate in $b_k$ will be low, i.e., the algorithm will be more certain (i.e., exploitation). When the shift happens at time step $\Bar{T}$, $p(s_{t+1}|s_{t}, a_{t};\theta)$ behaves as a out-of-distriubiton detector by its likelihood will measure the ratio of $\mathbb{P}/\bar{\mathbb{P}}$, decreasing value, yielding the exploration rate in $b_k$ will be high, i.e., the algorithm will be more uncertain (i.e., exploration).

\begin{figure}[t!]
\begin{algorithm}[H]
    \centering
    \caption{Our DQUCB algorithm in the infinite-horizon discounted MDP}\label{alg:infinite}
    \begin{algorithmic}[1]
        \footnotesize
        \STATE {\bfseries Initialize} $Q(s,a) \leftarrow \frac{1}{1-\gamma}$ \& $N(s,a)\leftarrow 0$  $\forall (s,a)\in \mathcal{S} \times \mathcal{A}$,  $p(\cdot;\theta)$, $\alpha_k=\frac{H+1}{H+k}$,  $\iota(k) = \log(|\mathcal{S}| |\mathcal{A}|T(k+1)(k+2))$, $H \leftarrow \frac{\ln(2/(1-\gamma))}{\ln(1/\gamma)}$
        \FOR{every step $t\in [T]$}
            \STATE Take $a_t\leftarrow \arg\max_{a'\in \mathcal{A}}Q(s_t,a')$, observe $s_{t+1}$
            \STATE $k=N(s_t,a_t) \leftarrow N(s_t,a_t)+1$
            \STATE $b_k \leftarrow \frac{c_2}{(1-\gamma)p(s_{t+1}|s_{t}, a_{t};\theta)}\sqrt{H \iota(k)/k}$
            \STATE $\hat{V}(s_{t+1}) \leftarrow \max_{a'\in \mathcal{A}}\hat{Q}(s_{t+1},a')$
            \STATE $Q(s_t,a_t) \leftarrow (1-\alpha_k) Q(s_t,a_t) + \alpha_k\left[r(s_t,a_t) + \gamma \hat{V}(s_{t+1}) + b_k\right]$
            \STATE $\hat{Q}(s_t,a_t) \leftarrow \min\left\{\hat{Q}(s_t,a_t), Q(s_t,a_t)\right\}$
            \STATE Update $\theta$ of model $p(\cdot;\theta)$ by $p(s_{t+1}|s_t, a_t;\theta)$
        \ENDFOR
    \end{algorithmic}
\end{algorithm}
\end{figure}

\section{Theoretical analysis}\label{sec:theorem}
To formally explain why our shift-aware DQUCB can improve the uncertainty estimation quality of UCB with Q-learning, resulting in a balance between exploration and exploitation under distribution shifts, we next analyze the cumulative regret of Alg.~\ref{alg:episodic} and Alg.~\ref{alg:infinite}. We first provide the regret bound for the Alg.~\ref{alg:episodic} in the finite-horizon episodic MDP under distribution shifts:
\begin{theorem}\label{thm:episode_bound}
   There exists an absolute constant $c>0$ such that for any $\delta \in (0,1)$, if we choose $b_t=\frac{c}{p(\cdot|s,a;\theta_h)}\sqrt{H^3\iota/t}$, where $t=N_h(s,a)$, then with probability $1-\delta$, the regret of the Algorithm~\ref{alg:episodic} satisfies
   \small
   \begin{align*}
        \text{Regret}(K) \leq \mathcal{O}\left(\sqrt{H^5(1+\epsilon)^2 |\mathcal{S}| |\mathcal{A}| K \log(|\mathcal{S}| |\mathcal{A}|KH/\delta)}\right),
    \end{align*}
    \normalsize
    where $\epsilon$ is the estimator error of $1/p(\cdot|s,a;\theta_h)$.
\end{theorem}

\begin{figure*}[t!]
    \centering
    \hspace*{-0.1in}
    \setlength{\tabcolsep}{0.1pt}
    \begin{tabular}{ccc}
    \includegraphics[width=0.33\linewidth]{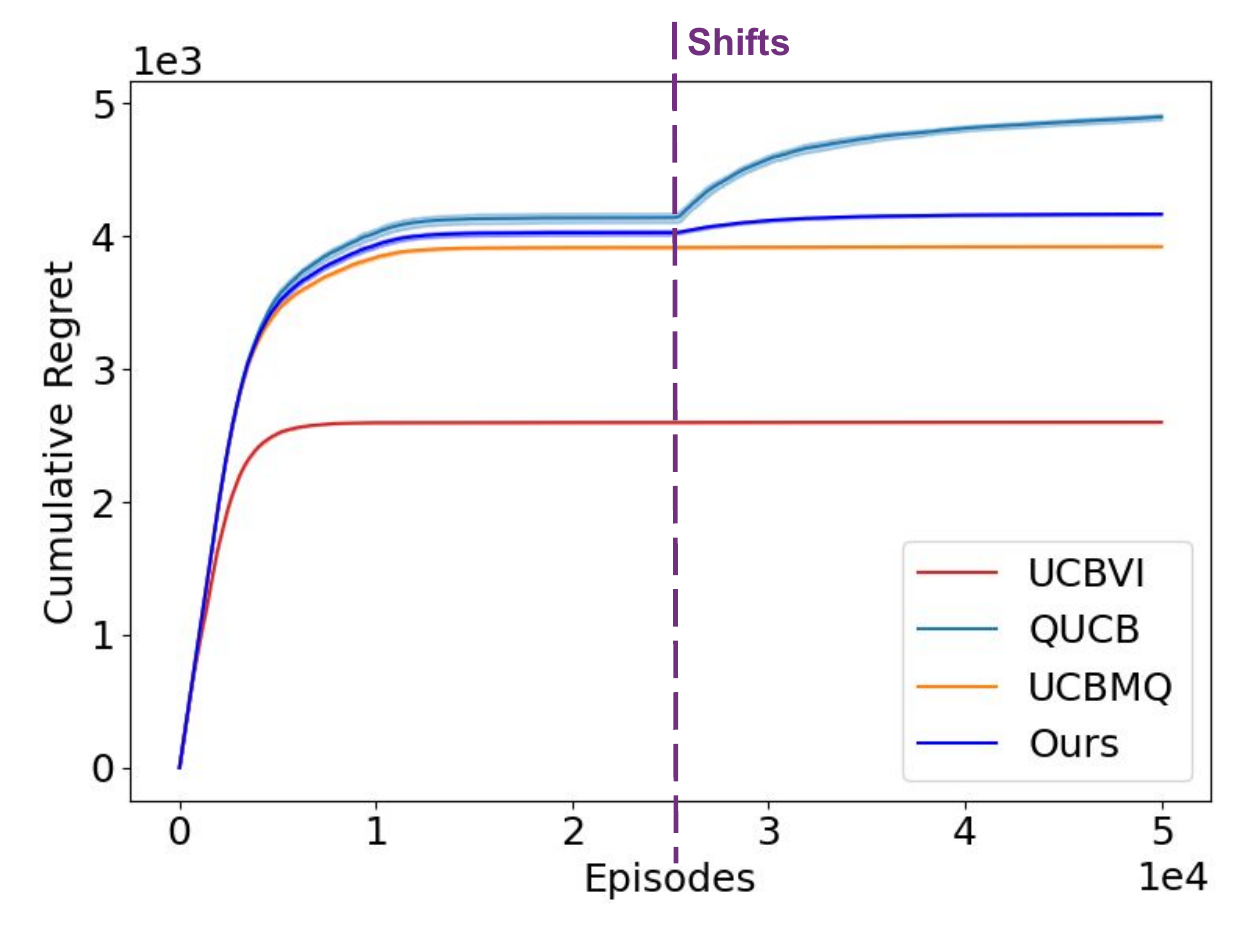}&
    \includegraphics[width=0.33\linewidth]{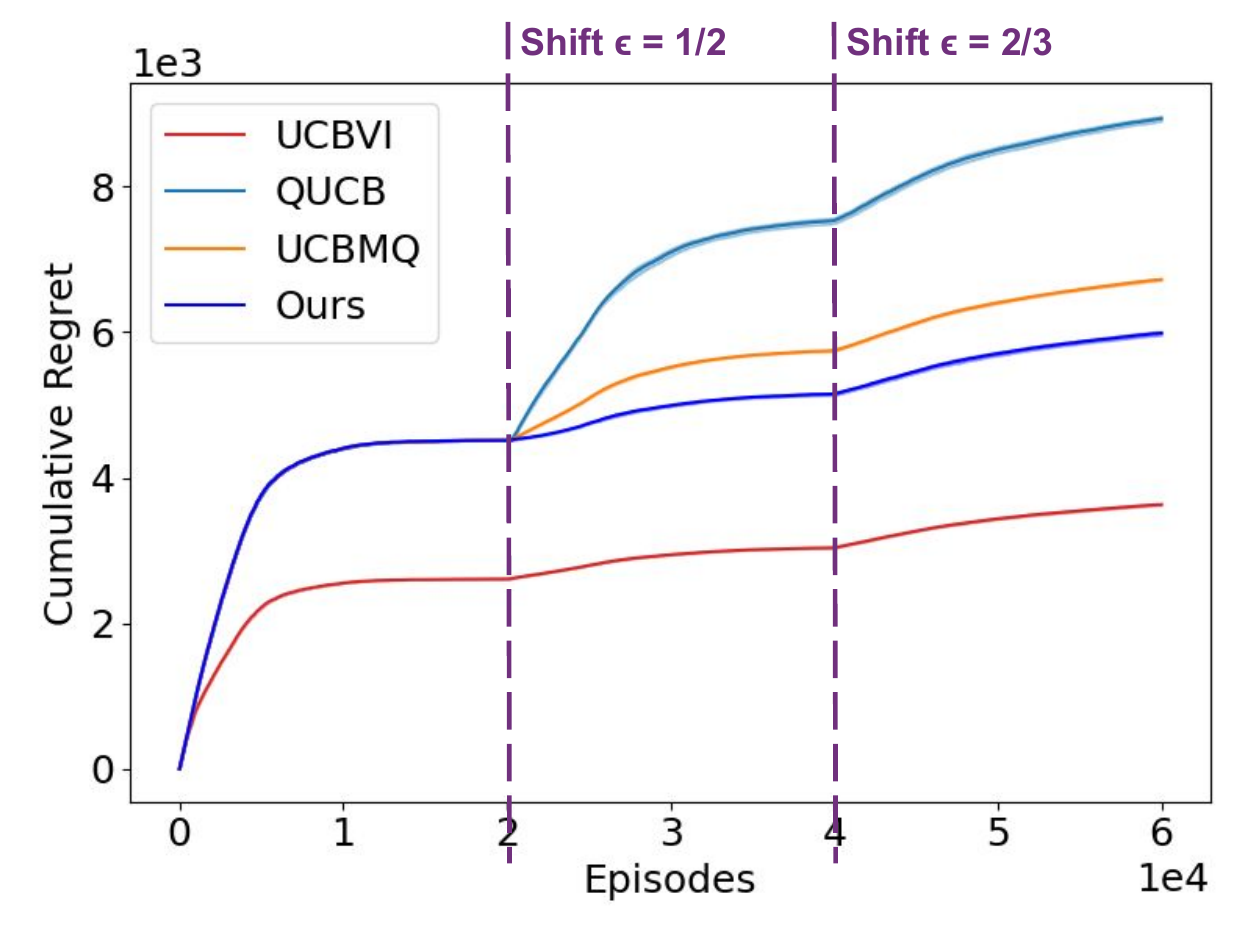}&
    \includegraphics[width=0.33\linewidth]{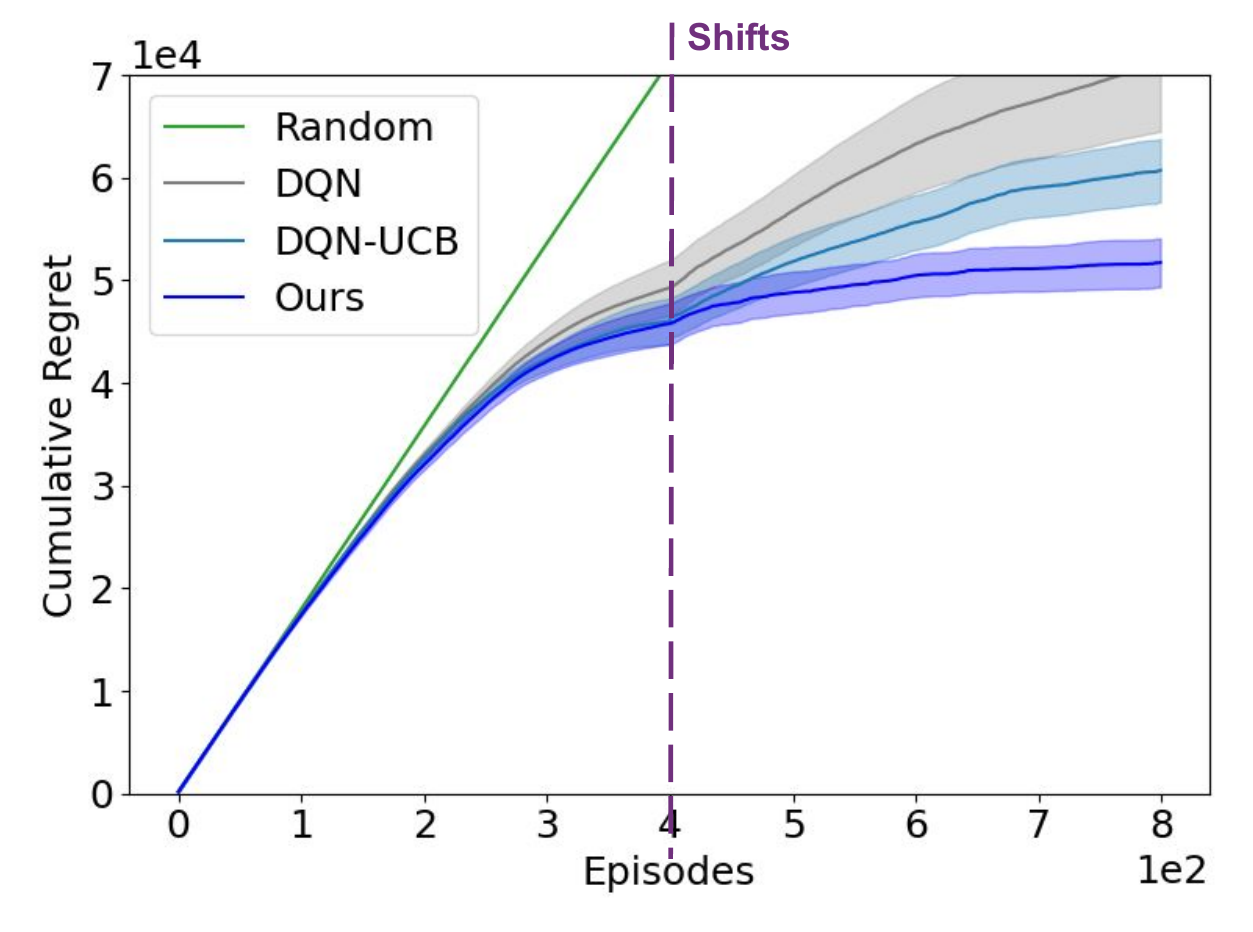}\\
    \small{(a)} & \small{(b)} & \small{(c)}
    \end{tabular}
    \caption{\small{(a) Cumulative regret on GridWorld, transition noise $\epsilon=0.01$ and $\epsilon=0.2$ before and after the shift at $\Bar{K}=25000$; (b) Cumulative regret on Frozen-Lake, slippery level $\epsilon=0$, $\epsilon=1/2$, and $\epsilon=2/3$ before and after the shift at $\Bar{K}=\{20000, 40000\}$; (c) Cumulative regret on CartPole, the transition noise $\mathcal{N}(0,0.15)$ is added to the velocity state after episode $\Bar{K}=400$. Results are average over $10$ runs. We refer to the computational complexity in Fig.~\ref{fig:complexity}.}}
    \label{fig:regret}
\end{figure*}

\begin{remark}\label{rmk:episode_bound}
    Thm.~\ref{thm:episode_bound} shows that when the estimation error $\epsilon \rightarrow 0$, our DQUCB in Alg.~\ref{alg:episodic} achieves $\mathcal{O}\left(\sqrt{H^5 |\mathcal{S}| |\mathcal{A}| K \log(|\mathcal{S}| |\mathcal{A}|KH/\delta)}\right)$ regret bound. Meanwhile, due to the distribution shift, the non-shift-aware version, i.e., Q-learning Hoeffding UCB~\citep{jin2018isQ} with the selection of $b_k=c\sqrt{H^3\iota/t}$, can leads to $\mathcal{O}\left(\sqrt{H^5 |\mathcal{S}| |\mathcal{A}| \bar{K} \log(|\mathcal{S}| |\mathcal{A}|\bar{K}H/\delta)} + H(K-\bar{K})\right)$. Since $\lim_{K\rightarrow\infty}\frac{\sqrt{H^5 |\mathcal{S}| |\mathcal{A}| K \log(|\mathcal{S}| |\mathcal{A}|KH/\delta)}}{\sqrt{H^5 |\mathcal{S}| |\mathcal{A}| \bar{K} \log(|\mathcal{S}| |\mathcal{A}|\bar{K}H/\delta)} + H(K-\bar{K})}=0$ by L'Hôpital's rule, we can see that the regret bound of our shift-aware DQUCB in  Alg.~\ref{alg:episodic} when $\epsilon \rightarrow 0$, is strictly better than the non-shift-aware version QUCB.
\end{remark}

It is worth noticing that in the finite-horizon episodic MDP, the regret in Thm.~\ref{thm:episode_bound} only needs to be measured at the initial state $s_1$, i.e., $(V_1^*-V_1^\pi)(s_1)$ and $(\bar{V}_1^*-\bar{V}_1^\pi)(s_1)$ in the first $KH$ steps. However, this analysis is non-trivial to extend to the infinite horizon setting because the agent may enter under-explored regions at any time $t\in [T]$~\citep{wang2020Qlearning}. Therefore, it is necessary to introduce the sub-optimality gap~\citep{yang2021qlearning}, i.e., $\Delta(s,a):=V^*(s)-Q^*(s,a)$ and $\bar{\Delta}(s,a):=\bar{V}^*(s)-\bar{Q}^*(s,a)$ to bound the gap at timestep $t$ and state $s_t$, i.e., $(V^*-V^{\pi_t})(s_t)$ and $(\bar{V}^*-\bar{V}^{\pi_t})(s_t)$. From this definition, we finally provide the regret bound for Alg.~\ref{alg:infinite} in the infinite-horizon discounted MDP under distribution shifts:
\begin{theorem}\label{thm:infiteMDP_bound}
   There exists an absolute constant $c_2>0$ such that for any $\delta \in (0,1)$, if we choose $b_k=\frac{c_2}{p(\cdot|s,a;\theta)}\sqrt{H\iota(k)/k}$, then with probability $1-\delta$,  the regret of the Algorithm~\ref{alg:infinite} satisfies
   \begin{align*}
        \text{Regret}(T) &\leq \mathcal{O}\left(\frac{|\mathcal{S}| |\mathcal{A}|(1+\epsilon)}{(1-\gamma)^6\min\{\Delta_{\min}, \bar{\Delta}_{\min}\}}\right.\\
        &\quad \left. \cdot \log\left(\frac{|\mathcal{S}| |\mathcal{A}|T}{(1-\gamma)\min\{\Delta_{\min}, \bar{\Delta}_{\min}\}}\right)\right),
    \end{align*}
    where $\epsilon$ is the estimator error of $1/p(\cdot|s,a;\theta)$, $\Delta_{\min}:=\min_{(s,a)\in\mathcal{S}\times \mathcal{A}}\{\Delta(s,a):\Delta(s,a)\neq 0\}$, and $\bar{\Delta}_{\min}:=\min_{(s,a)\in\mathcal{S}\times \mathcal{A}}\{\bar{\Delta}(s,a):\bar{\Delta}(s,a)\neq 0\}$.
\end{theorem}
The proof for Thm.~\ref{thm:episode_bound} and Thm.~\ref{thm:infiteMDP_bound} decompose the regret to bounding the learning error before and after the shift at episode $\bar{K}$ and time step $\bar{T}$, use the density function to construct the concentration bound of $Q$-value with Lem.~\ref{lemma:boundQ} and Lem.~\ref{lemma:boundQ2}, and finally adapt techniques of the learning error recursion from~\citep{jin2018isQ,wang2020Qlearning}, details are in Apd.~\ref{apd:thm:episode_bound} and Apd.~\ref{apd:thm:infiteMDP_bound}, respectively.

\begin{remark}
    Thm.~\ref{thm:infiteMDP_bound} shows that when the estimation error $\epsilon \rightarrow 0$, our DQUCB in Alg.~\ref{alg:infinite} achieves $\mathcal{O}\left(\frac{|\mathcal{S}| |\mathcal{A}|}{(1-\gamma)^6\min\{\Delta_{\min}, \bar{\Delta}_{\min}\}} \log\left(\frac{|\mathcal{S}| |\mathcal{A}|T}{(1-\gamma)\min\{\Delta_{\min}, \bar{\Delta}_{\min}\}}\right)\right)$ regret bound. Meanwhile, due to the distribution shift, the non-shift-aware version, i.e., Q-learning Hoeffding UCB~\citep{wang2020Qlearning} with the selection of $b_k=c_2\sqrt{H\iota(k)/k}$, can leads to $\mathcal{O}\left(\frac{|\mathcal{S}| |\mathcal{A}|}{(1-\gamma)^6\Delta_{\min}} \log\left(\frac{|\mathcal{S}| |\mathcal{A}|\bar{T}}{(1-\gamma)\Delta_{\min}}\right)\right) + (T-\bar{T})$. Similar to the limit of a function analysis in Remark~\ref{rmk:episode_bound}, when $T\rightarrow \infty$, we can see that the regret bound of our shift-aware DQUCB in Alg.~\ref{alg:infinite} is strictly better than the non-shift-aware version QUCB.
\end{remark}

\textbf{Justification on how density estimator error $\epsilon$ scales with time or sample}. The density estimator error $\epsilon$ depends on the specific density estimation methods. In our experiment, we use the Kernel Density Estimation, hence, the $\epsilon=\mathcal{O}(n_t^{-4/5})$, where $n_t$ is the number of cumulative data points at time $t$. For other parametric methods, the error can be reduced to $\epsilon=\mathcal{O}(n_t^{-1})$~\citep{wahba1975optimal}. In general, over time, the more samples we have, the smaller the estimator error bound, leading to a smaller gap between our oracle and practical algorithm.
\section{Experiments}
\subsection{Experimental settings}
We illustrate the benefits of our DQUCB across four different tasks. In the first two tasks, we compare with other tabular UCB-based RL baselines (i.e., QUCB~\citep{jin2018isQ}, UCBMQ~\citep{menard2021UCBmomentum}, and UCVI~\citep{azar2017minimax}) in the GridWorld and Frozen Lake environments. In the third task, we extend our DQUCB to an infinite-state with Deep-RL on CartPole-v0~\citep{brockman2016gym}. In the final task, we extend this to a real-world healthcare application. We optimize KDE with the most recent $n=100$ cumulative samples. In the Deep-RL setting, since the UCB term needs to store the number of visited pairs $N_h(s,a)$, we adapt the hashing technique of~\citet{tang2017countdeep} to count this number. In GridWorld and Frozen Lake, the shift is the change of the slippery level when moving between states. In CartPole, the shift is the change of the Gaussian noise level to the cart and pole angular velocity features. In the healthcare dataset, the shift is the change of the parameter for updating the estimated COVID-19 occupancy. More details are in Apd.~\ref{apd:exp_settings}.

\subsection{Main results}
\subsubsection{GridWorld and Frozen Lake}
From Fig.~\ref{fig:regret}~(a) and Fig.~\ref{fig:regret}~(b), we observe that our shift-aware DQUCB (colored by dark blue) outperforms the non-shift-aware version (i.e., QUCB) by having a lower cumulative regret across episodes. Notably, we can see that due to being able to detect the distribution shift, our method can explore new optimal policies, resulting in significantly lower regret than QUCB after the shift happens. This result confirms our theoretical guarantee in Sec.\ref{sec:theorem}.

\begin{figure}[ht!]
    \centering
    \includegraphics[width=1.0\linewidth]{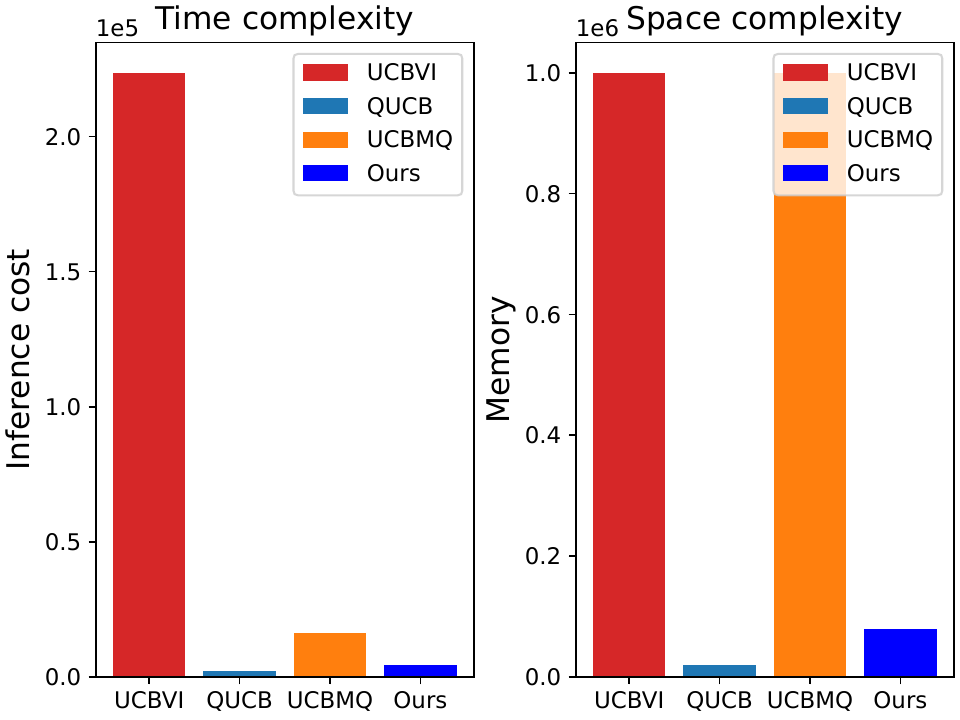}
    \caption{\small{Time and space complexity comparison (lower are better) in the $[10]\times[5]$ GridWorld environment.}}
    \label{fig:complexity}
\end{figure}

It is also worth noticing that while having a higher regret than the model-based RL (i.e., UCVI) and competitive results with the mixture of model-based and model-free RL (i.e., UCBMQ), \textbf{our method is much computationally efficient with a remarkably lower time and space complexity in Fig.~\ref{fig:complexity}}. This is because UCVI and UCBMQ need to build value functions for each state-action pair, and need to iterate through all possible $(s',s,a) \in \mathcal{S} \times \mathcal{S} \times \mathcal{A}$ tuples per each time step (see Remark~\ref{rmk:complexity}). In contrast, our method only needs to optimize the density function with cumulative observed tuples. Therefore, these results suggest that our method can balance between regret and computational efficiency, and can be deployed in complex environments or low-resource applications in the real world. More results and analysis are provided in Apd.~\ref{apd:add_results}.

\subsubsection{CartPole}

\begin{figure}[ht!]
    \centering
    \includegraphics[width=1.0\linewidth]{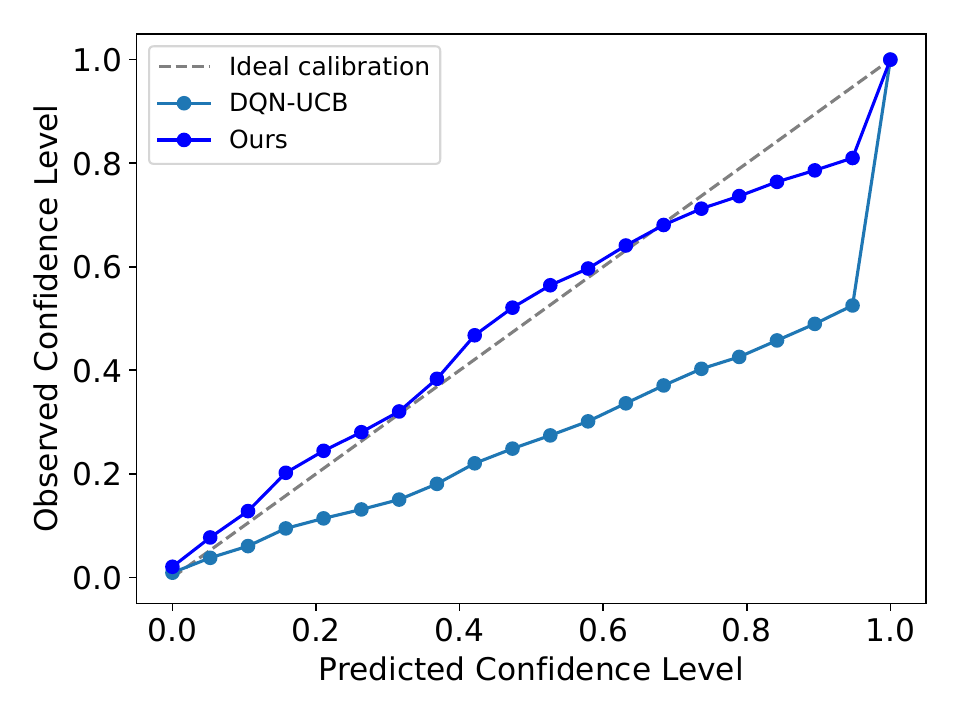}
    \caption{\small{Uncertainty quantification quality of UCB with calibration error measurement.}}
    \label{fig:calib}
\end{figure}

\begin{figure*}[t!]
    \centering
    \hspace*{-0.1in}
    \setlength{\tabcolsep}{0.1pt}
    \begin{tabular}{ccc}
    \includegraphics[width=0.33\linewidth]{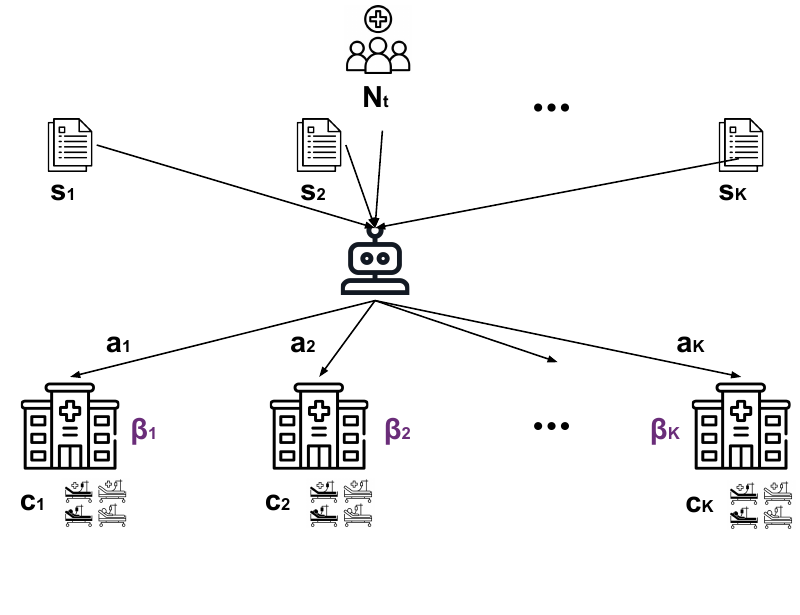}&
    \includegraphics[width=0.33\linewidth]{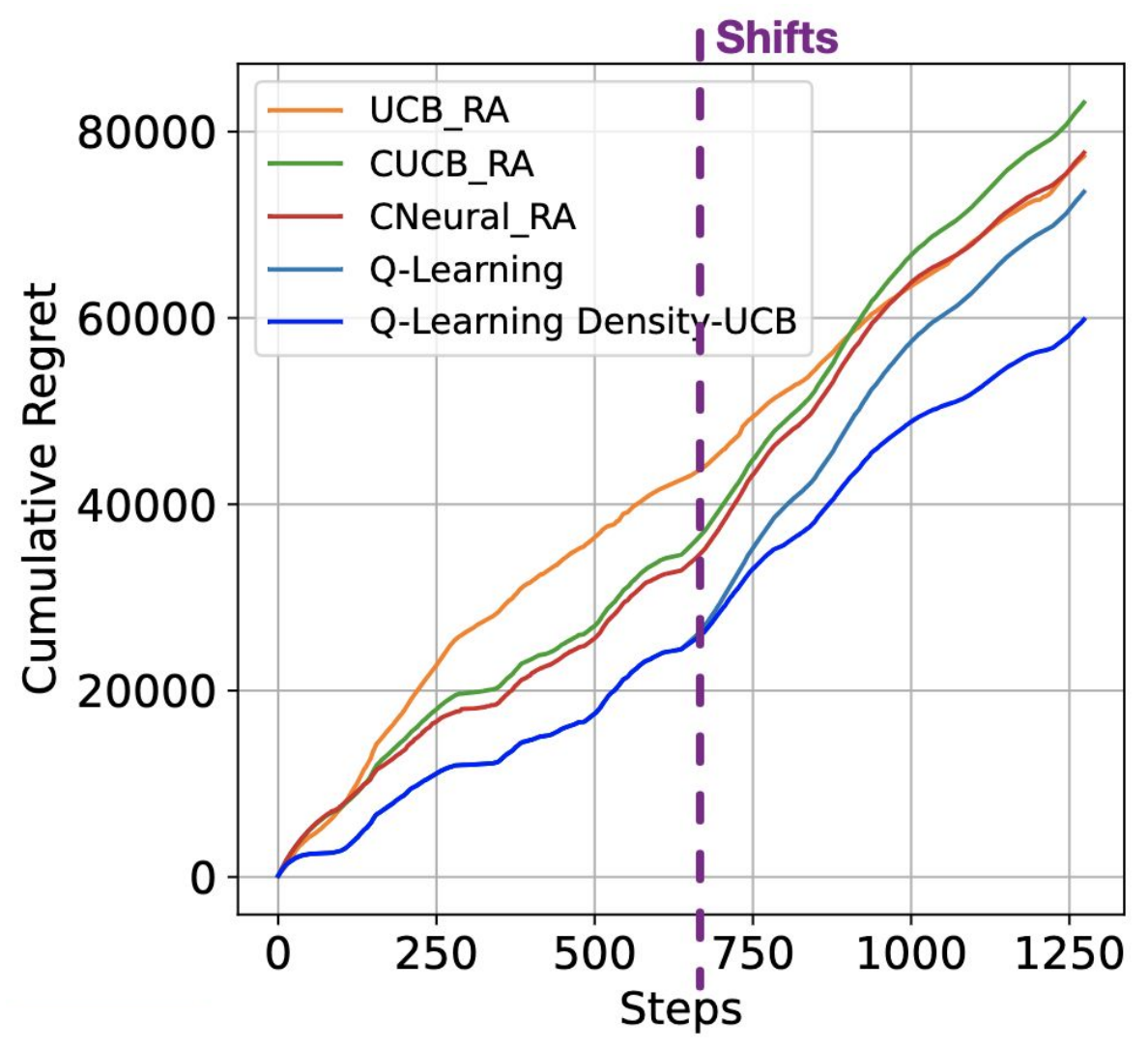}&
    \includegraphics[width=0.33\linewidth]{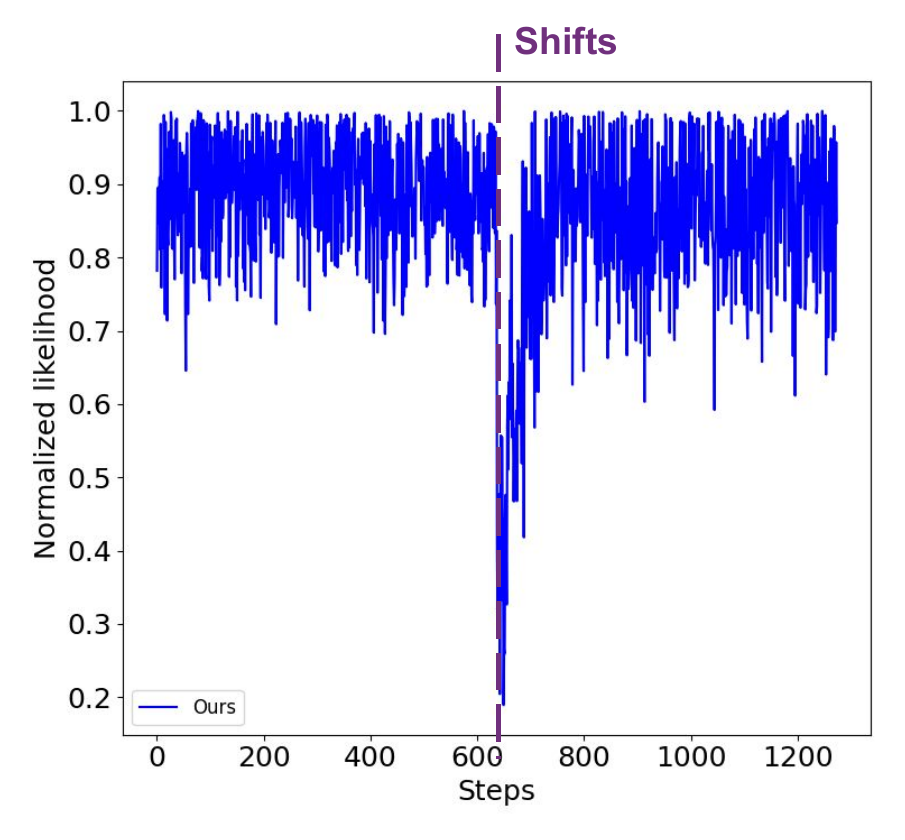}\\
    \small{(a)} & \small{(b)} & \small{(c)}
    \end{tabular}
    \caption{\small{Infinite-horizon discounted MDP setting: (a) Agent receives $K$ hospital status states, needs to allocate $N_t$ patients to $K$ hospitals s.t. minimize the number of overflow; (b) Performance across $K=40$ Texas hospitals daily from 2020-08 to 2024-01; (c) Normalized likelihood value of our density function across time steps.}}
    \label{fig:healthcare}
\end{figure*}

To illustrate the merits of our method in complex domains, we deploy it in an infinite-state environment with CartPole and Deep-RL. Since the state space is infinite, UCVI and UCBMQ are non-trivial in extending to this setting because they need to iterate through all possible state-action pairs. Hence, we compare our method with 3 baselines, including random policy, Deep Q-Network (DQN)~\citep{paszke2019torch}, and its UCB extension, i.e., DQN-UCB. Although training Q-learning with neural networks is quite unstable, Fig.~\ref{fig:regret}~(c) still shows that our method achieves the lowest cumulative regret, especially under the distribution shift. This once again confirms our tighter regret bound in Sec.~\ref{sec:theorem}. To better understand the UCB quality with Q-values, Fig.~\ref{fig:calib} evaluates the calibration performance. Intuitively, calibration means a $p$ confidence interval contains the reward $p$ of the time~\citep{bui2025varianceaware}. We can see that by being unable to detect the distribution shift, DQN-UCB is over-confident in its prediction. Meanwhile, by leveraging the density function to detect the shift, our method is well-calibrated and significantly better than DQN-UCB.

\subsection{COVID-19 patient hospital allocation}
Finally, we evaluate our model's performance on a COVID-19 patient hospital allocation. Specifically, given $K$ hospitals in Fig.~\ref{fig:healthcare}~(a), the agent receives the state $s_t\in \mathbb{R}^{2K+1}$ of a hospital at time $t$. The state $s_t=(y_{1,t}, \cdots, y_{K,t}, d_{1,t}, \cdots, d_{K,t}, N_t)$, where $y_{i,t}$ and $d_{i,t}$ are the number of COVID-19 patient occupancy and non-COVID patients in hospital $i$ on day $t$, and $N_t$ is the number of arriving COVID-19 patients in the system on day $t$. For every day $t$, the agent must allocate these $ N_t$ patients to $K$ hospitals by action $a_t=(a_{1,t},\cdots, a_{K,t})$ to minimize the number of overflows across hospitals, i.e., maximize the reward $r_t=\sum_{i=1}^K -\max\left(0,(\beta_i \cdot y_{i,t} + a_{i,t+1} + d_{i,t+1}) - c_i\right)$, where $c_i$ is the capacity of the hospital $i$ and $\beta_i$ is the parameter provided by the environment to update the estimated occupancy of COVID-19 (details are in Apd.~\ref{apd:exp_settings}). We use the environment parameter $\beta$ and the number of arriving COVID-19 patients $N_t$ from the real-world dataset: COVID-19 Reported Patient Impact \& Hospital Capacity by Facility, provided by the U.S. Department of Health \& Human Services over $T=1274$ days from 2020 to 2024. The environment is shifted by changing $\beta$ to be uniform at day $\Bar{T} = 637$. We use a neural-net to estimate the $Q_t$ matrix, where $Q[i,j]$ means how many people can be saved if we allocate $j$ patients to hospital $i$. Then, allocate $N_t$ patients to the hospital with action $a_t$ by using an oracle from Q-values~\citep{zuo2021combinatorialmultiarmedbanditsresource,li2024deepreinforcementlearningefficient}.

Fig.~\ref{fig:healthcare}~(b) summarizes our results with the cumulative regret (optimal rewards are obtained from the dataset). We observe that before the shifts happen, our method and DQN-UCB outperform other baselines~\citep{zuo2021combinatorialmultiarmedbanditsresource} with the lowest cumulative regret. Notably, when the shift occurs, our density function decreases the likelihood value in Fig.~\ref{fig:healthcare}~(c), signaling to the model that the environment is changing. This yields our UCB value increase, i.e., the model increases its uncertainty and encourages exploring new policies. As a result, compared to the non-shift-aware DQN-UCB method, our method achieves better adaptation with significantly lower regret after the shift occurs. This result suggests that our method can allocate COVID-19 patients to hospitals better than other baselines, so that we can save more people in the real world.
\section{Related work}
\textbf{UCB-based methods in non-stationary RL.} Relying on the principle of optimism in the face of uncertainty, UCB has become an efficient strategy to explore the uncertain environment while maximizing the reward in bandit~\citep{yasin2011improved}. Building on this direction, a recent line of research has applied these UCB ideas to address non-stationary RL, encompassing both model-based and model-free settings. Specifically, in the model-based RL, UCLR2~\citep{auerUCLR2} and UCLR3~\citep{bourel2020UCLR3} form estimates of the transition probabilities of the MDP using past samples, and add UCB to the estimated transition matrix. An alternative approach is UCBVI~\citep{azar2017minimax}, which directly adds a UCB bonus to the Q-values with a strong assumption that transition matrices are similar within an episode. Despite the guarantees of sharp regret and the match with the lower bound (that is, $\Omega(\sqrt{H^3|\mathcal{S}| |\mathcal{A}|K})$), all of the results in this model-based research require estimating and storing the entire transition matrix and thus suffer from unfavorable time (i.e., $\Tilde{\mathcal{O}}(KH|\mathcal{S}|^2|\mathcal{A}|)$) and space complexities (i.e., $\mathcal{O}(|\mathcal{S}|^2|\mathcal{A}|H)$).

Therefore, toward a computationally efficient method, \citet{jin2018isQ} introduces model-free Q-learning~UCB by incorporating an UCB term to show the confidence of $Q$-values, which can achieve total regret $\mathcal{O}(\sqrt{H^5|\mathcal{S}||\mathcal{A}|K})$ with $\mathcal{O}(KH)$ time and $\mathcal{O}(|\mathcal{S}||\mathcal{A}|H)$ space complexity in episodic MDP. Based on this direction, \citet{wang2020Qlearning} and \citet{yang2021qlearning} later show that Q-learning~UCB achieves nearly optimal regret bound in discounted MDP. In particular, if there exists a strictly positive sub-optimality gap, \citet{wang2020Qlearning} proves that Q-learning enjoys a $\mathcal{O}\left(\frac{|\mathcal{S}| |\mathcal{A}|}{(1-\gamma)^6\Delta_{\min}} \log\left(\frac{|\mathcal{S}| |\mathcal{A}|T}{(1-\gamma)\Delta_{\min}}\right)\right)$ cumulative regret bound. Yet, all of these results for Q-learning~UCB assume that the transition between episodes in episodic MDP and the step in discounted MDP are fixed. This is a strong assumption in real-world sequential decision-making applications because the environment can change in the long run. Hence, we relax this fixed assumption by considering that the transition function can change at a particular episode and time step.

There are also other variants of non-stationary RL settings, such as broader literature on continual or concept-drift RL~\citep{cheung2020RL, xie2021deepRL}. Some close to our setting, like \citet{gajane2018sliding}, present a non-UCB approach that considers a switching-MDP problem, rather than our non-stationary problem. \citet{mao2021near} constrains the transition cumulative variations to not exceed certain variation budgets, and the regret bound depends on this budget value, whereas our setting and analysis do not depend on any budget.
 
\textbf{Improving UCB quality}. Using a density function to improve model uncertainty has been studied in classification and regression~\citep{bui2024densifysoftmax,bui2024density}, but only in supervised learning. In sequential decision problems, improving the UCB quality to obtain better regret has shown promising results in bandit~\citep{auer2002finitetime,kuleshov2014algorithms,zhou2021nearly}. For example, ~\citet{calibrated2019malik,deshpande2024online} have empirically shown that calibrated UCB algorithms can result in lower cumulative regret. Theoretically, \citep{zhao2023variance,bui2025varianceaware} show that variance-aware~UCB, i.e., using the reward noise variance to improve UCB quality, can further achieve a tighter regret bound. Regarding RL, \citet{strehl2006delayedQ} introduced delayed Q-learning, where the $Q$-value for each state-action pair is updated only once every certain number of times this pair is visited. Yet, it is quite sample-inefficient compared to other approaches~\citep{jin2018isQ}. Closest to our work is UCBMQ~\citep{menard2021UCBmomentum,bowen2021finite,azar2011speedQ}, which is also based on the Q-learning UCB, but additionally incorporates a momentum term that is built from the value functions for each state-action pair. Because of needing to compute these bias-value functions for every round, UCBMQ is significantly less computationally efficient than model-free baselines by its $\mathcal{O}(H(|\mathcal{S}| + |\mathcal{A}|)K)$ time and $\mathcal{O}(|\mathcal{S}|^2|\mathcal{A}|H)$ space complexity. Beyond this computational limitation, UCBMQ is also designed only for tabular episodic MDPs. Meanwhile, our DQUCB is more computationally efficient, designed for both episodic and discounted MDPs, and can extend to Deep-RL with a continuous state-action space.
\section{Conclusion and Discussion}
Q-learning with UCB exploration has become a standard model-free RL that is provably efficient, with nearly optimal regret bound in non-stationary RL. Yet, it can exploit sub-optimal reward if the environment suddenly changes in some episode in the finite-horizon episodic MDP, or step in the infinite-horizon discounted MDP. To tackle this challenge, we introduce DQUCB, a shift-aware Q-learning~UCB that leverages the transition density function to enhance the uncertainty quantification of the UCB, resulting in better exploration and exploitation. Our theoretical results show that our oracle DQUCB can achieve a nearly optimal regret bound in both episodic MDP and discounted MDP under distribution shifts. Our empirical results demonstrate the computational efficiency of our method when compared to model-based baselines and confirm our theoretical analysis with a significantly lower regret than Q-learning baselines across different tasks, datasets, and model architectures. With these results, we hope our work will contribute to the literature on improving the uncertainty and robustness of UCB for Q-learning in sequential problems. 

Our proposed method is based on Q-learning, which is specifically designed for discrete action spaces, and requires conversion between continuous and discrete spaces for applications in continuous spaces. In addition, the quality of our UCB relies on the quality of the density function; if the density estimator is not sufficiently accurate, it can affect the performance of our model. Future work includes tackling the limitation of density estimation with the kernel estimator and extending our method to more RL tasks.

\section*{Acknowledgement}
This work is supported by a seed grant from JHU Institute of Assured Autonomy.

\bibliography{refs}
\bibliographystyle{plainnat}

\section*{Checklist}

\begin{enumerate}

  \item For all models and algorithms presented, check if you include:
  \begin{enumerate}
    \item A clear description of the mathematical setting, assumptions, algorithm, and/or model. [Yes]
    \item An analysis of the properties and complexity (time, space, sample size) of any algorithm. [Yes]
    \item (Optional) Anonymized source code, with specification of all dependencies, including external libraries. [Yes]
  \end{enumerate}

  \item For any theoretical claim, check if you include:
  \begin{enumerate}
    \item Statements of the full set of assumptions of all theoretical results. [Yes]
    \item Complete proofs of all theoretical results. [Yes]
    \item Clear explanations of any assumptions. [Yes]     
  \end{enumerate}

  \item For all figures and tables that present empirical results, check if you include:
  \begin{enumerate}
    \item The code, data, and instructions needed to reproduce the main experimental results (either in the supplemental material or as a URL). [Yes]
    \item All the training details (e.g., data splits, hyperparameters, how they were chosen). [Yes]
    \item A clear definition of the specific measure or statistics and error bars (e.g., with respect to the random seed after running experiments multiple times). [Yes]
    \item A description of the computing infrastructure used. (e.g., type of GPUs, internal cluster, or cloud provider). [Yes]
  \end{enumerate}

  \item If you are using existing assets (e.g., code, data, models) or curating/releasing new assets, check if you include:
  \begin{enumerate}
    \item Citations of the creator If your work uses existing assets. [Yes]
    \item The license information of the assets, if applicable. [Not Applicable]
    \item New assets either in the supplemental material or as a URL, if applicable. [Not Applicable]
    \item Information about consent from data providers/curators. [Not Applicable]
    \item Discussion of sensible content if applicable, e.g., personally identifiable information or offensive content. [Not Applicable]
  \end{enumerate}

  \item If you used crowdsourcing or conducted research with human subjects, check if you include:
  \begin{enumerate}
    \item The full text of instructions given to participants and screenshots. [Not Applicable]
    \item Descriptions of potential participant risks, with links to Institutional Review Board (IRB) approvals if applicable. [Not Applicable]
    \item The estimated hourly wage paid to participants and the total amount spent on participant compensation. [Not Applicable]
  \end{enumerate}

\end{enumerate}

\newpage
\appendix
\onecolumn
\aistatstitle{Q-Learning with Shift-Aware Upper Confidence Bound in Non-Stationary Reinforcement Learning (Supplementary Material)}

In this supplementary material, we collect proofs and remaining materials deferred from the main paper. In Appendix~\ref{apd:proof}, we provide the proofs for all our theoretical results, including: proof of Theorem~\ref{thm:episode_bound} in Appendix~\ref{apd:thm:episode_bound}; proof of Theorem~\ref{thm:infiteMDP_bound} in Appendix~\ref{apd:thm:infiteMDP_bound}; proof of Lemma~\ref{lemma:boundQ} in Appendix~\ref{proof:lemma:boundQ}; proof of Lemma~\ref{lemma:boundQ2} in Appendix~\ref{proof:lemma:boundQ2}. In Appendix~\ref{apd:exp}, we provide additional information about our experiments, including: sufficient details about experimental settings in Appendix~\ref{apd:exp_settings}; demo code in Appendix~\ref{apd:code}; additional results in Appendix~\ref{apd:add_results} with performance across different shift intensities, types of environment shift, further ablation studies for comparison with model-based RL, density estimation, and extension to continuous action space experiments. Finally, the source code to reproduce our results is available at \href{https://github.com/Angie-Lab-JHU/density-Qlearning-UCB}{https://github.com/Angie-Lab-JHU/density-Qlearning-UCB}.

\section{Proofs}\label{apd:proof}
In this section, we first formally introduce notations that will be used intensively in our proofs in Appendix~\ref{apd:notation}, and then summarize our useful lemmas to support our proofs in Appendix~\ref{apd:useful_lemmas}.

\subsection{Notations}\label{apd:notation}
\textbf{Notation in the finite-horizon episodic MDP}. Let $\mathbb{I}[A]$ as the indicator function for event $A$, $(s_h^k, a_h^k)$ as the actual state-action pair observed and chosen at step $h$ of episode $k$, and $Q_h^k, V_h^k, N_h^k$ as the $Q_h,V_h,N_h$ functions at the beginning of episode $k$, respectively. Using this notation and let $t=N_h^k(s,a)$ and suppose $(s,a)$ was previously taken at step $h$ of episodes $k_1,\cdots, k_t<k$, the update equation at episode $k$ in Algorithm~\ref{alg:episodic} can be rewritten as follows, for every $h\in [H]$,
\begin{align}\label{eq:empiricalQ}
    &Q_{h}^k(s,a) = \alpha_t^0 H + \sum_{i=1}^t \alpha_t^i \left[r_h(s,a) + V_{h+1}^{k_i}(s_{h+1}^{k_i}) + b_i\right],\\
    &V_h^k(s) \leftarrow \min\{H,\max_{a'\in \mathcal{A}}Q_{h}^k(s,a')\}, \forall s\in \mathcal{S}.
\end{align}
Let $\left[\mathbb{P}_h V_{h+1}\right](s,a) := \mathbb{E}_{s'\sim \mathbb{P}_h(\cdot|s,a)}V_{h+1}(s')$ and its empirical counterpart of episode $k$ as $\left[\hat{\mathbb{P}}_h V_{h+1}\right](s,a) := V_{h+1}(s_{h+1})$, which is defined only for $(s,a)=(s_h^k, a_h^k)$.

\textbf{Notation in the infinite-horizon discounted MDP}. Let $Q^t$,  $\hat{Q}^t$, $V^t$, $\hat{V}^t$, $N^{t}$ are the value of $Q$, $\hat{Q}$, $V$, $\hat{V}$, $N$ right before the $t$-th step, respectively. Let $\tau(s,a,i)=\max\{t:N^t(s,a)=i-1\}$ be the step $t$ at which $(s^t,a^t)=(s,a)$ for the $i$-th time. From the update rule in the Algorithm~\ref{alg:infinite}, we have
\begin{align}\label{eq:empiricalQ2}
    &\hat{Q}^t(s,a) = \alpha_t^0 \frac{1}{1-\gamma} + \sum_{i=1}^t \alpha_t^i \left[r(s,a) + \gamma \hat{V}_{t+i}(s_{t_i+1}) + b_t\right],\\
    &Q^t(s,a) = \min\{\hat{Q}^t(s,a), Q^t(s,a)\}, \quad \hat{V}^t(s) \leftarrow \max_{a'\in \mathcal{A}}\hat{Q}^t(s,a'), \forall s\in \mathcal{S}.
\end{align}

\subsection{Useful Lemmas}\label{apd:useful_lemmas}
Our proofs in this section are based on the following provable lemmas:

\begin{lemma}(Learning rate properties~\citep{jin2018isQ,wang2020Qlearning})\label{lemma:learningrate}.
    For the learning rate $\alpha_t=\frac{H+1}{H+t}$, let $\alpha_t^0=\prod_{j=1}^{t}(1-\alpha_j)$, $\alpha_t^i=\alpha_i \prod_{j=i+1}^{t}(1-\alpha_j)$, then we have $\sum_{i=1}^t \alpha_t^i = 1$ and $\alpha_t^0=0$ for $t\geq 1$; $\sum_{i=1}^t \alpha_t^i=0$ and $\alpha_t^0=1$ for $t=0$. And, the following properties hold for $\alpha_t^i$:
    \begin{enumerate}[label=(\alph*),itemsep=1pt,topsep=0pt,parsep=0pt,leftmargin=*]
        \item $\frac{1}{\sqrt{t}}\leq \sum_{i=1}^t \frac{\alpha_t^i}{\sqrt{i}}\leq \frac{2}{\sqrt{t}}$ for every $t\geq 1$.
        \item $\max_{i\in [t]} \alpha_t^i \leq \frac{2H}{t}$ and $\sum_{i=1}^t (\alpha_t^i)^2 \leq \frac{2H}{t}$ for every $t\geq 1$.
        \item $\sum_{t=i}^\infty \alpha_t^i = 1+\frac{1}{H}$ for every $i\geq 1$.
        \item $\sqrt{\frac{\iota(t)}{t}}\leq \sum_{i=1}^t \alpha_t^i \sqrt{\frac{\iota(i)}{i}}\leq 2\sqrt{\frac{\iota(t)}{t}}$, where $\iota(t)=\ln(c(t+1)(t+2))$, for every $t\geq 1$, $c\geq 1$.
    \end{enumerate}
\end{lemma}

\begin{lemma}(Bound on $Q^k_h-Q^*$ and $Q^k_h-\bar{Q}^*$).\label{lemma:boundQ} There exists an absolute constant $c>0$ s.t., for any $p\in (0,1)$, letting $b_t = \frac{c}{p(\cdot|s,a;\theta_h)}\sqrt{H^3\iota/t}$ with $t=N_h(s,a)$, we have $\beta_t=2\sum_{i=1}^t \alpha_t^ib_i\leq 4\frac{c}{p(\cdot|s,a;\theta_h)}\sqrt{H^3\iota/t}$ and, for simultaneously $\forall (s,a,h,k)\in \mathcal{S}\times \mathcal{A}\times [H]\times [K]$,  with probability at least $1-\delta$, it holds that
\begin{align*}
    0\leq (Q_h^k-Q_h^*)(s,a)\leq \alpha_t^0 H +\sum_{i=1}^t \alpha_t^i \left(V_{h+1}^{k_i} - V_{h+1}^*\right)(s_{h+1}^{k_i}) + \beta_t,
\end{align*}
and
\begin{align*}
    0\leq (Q_h^k-\bar{Q}_h^*)(s,a)\leq \alpha_t^0 H +\sum_{i=1}^t \alpha_t^i \left(V_{h+1}^{k_i} - \bar{V}_{h+1}^*\right)(s_{h+1}^{k_i}) + \beta_t.
\end{align*}
The proof is in Appendix~\ref{proof:lemma:boundQ}.
\end{lemma}

\begin{lemma}(Bound on $Q^t-Q^*$ and $Q^t-\bar{Q}^*$).\label{lemma:boundQ2} There exists an absolute constant $c_2>0$ s.t., for any $p\in (0,1)$, letting $t=N_p(s,a)$, $t_i=\tau(s,a,i)$, $b_t = \frac{c_2}{(1-\gamma)p(\cdot|s,a;\theta)}\sqrt{H\iota(t)/t}$, we have $\beta_t\leq \frac{c_3}{(1-\gamma)p(\cdot|s,a;\theta)}\sqrt{H\iota(t)/t}$ and, for simultaneously $\forall (s,a,t)\in \mathcal{S}\times \mathcal{A}\times [T]$,  with probability at least $1-\delta$, it holds that
\begin{align*}
    0\leq (\hat{Q}^t-Q^*)(s,a)\leq (Q^t-Q^*)(s,a)\leq \frac{\alpha_t^0}{1-\gamma} +\sum_{i=1}^t \gamma \alpha_t^i \left(\hat{V}_{t_i} - V^*\right)(s_{t_i+1}) + \beta_t,
\end{align*}
and
\begin{align*}
    0\leq (\hat{Q}^t-\bar{Q}^*)(s,a)\leq (Q^t-\bar{Q}^*)(s,a)\leq \frac{\alpha_t^0}{1-\gamma} +\sum_{i=1}^t \gamma \alpha_t^i \left(\hat{V}_{t_i} - \bar{V}^*\right)(s_{t_i+1}) + \beta_t.
\end{align*}
The proof is in Appendix~\ref{proof:lemma:boundQ2}.
\end{lemma}

\subsection{Proof of Theorem~\ref{thm:episode_bound}}\label{apd:thm:episode_bound}
\begin{proof}
Recall by our notation in Appendix~\ref{apd:notation}, $\left[\mathbb{P}_h V_{h+1}\right](s,a) = \mathbb{E}_{s'\sim \mathbb{P}_h(\cdot|s,a)}V_{h+1}(s')$ and its empirical counterpart of episode $k$ is $\left[\hat{\mathbb{P}}_h V_{h+1}\right](s,a) = V_{h+1}(s_{h+1})$. By Lemma~\ref{lemma:boundQ}, with probability at least $1-\delta$, we have $Q_h^k\geq Q_h^*$ and $Q_h^k\geq \bar{Q}_h^*$, i.e., $V_h^k\geq V_h^*$ and $V_h^k\geq \bar{V}_h^*$, thus
\begin{align}\label{eq:origin_regret}
    \text{Regret}(K) &= \sum_{k=1}^{\bar{K}} (V_1^* - V_1^{\pi_k})(s_1^k) + \sum_{k=\bar{K}+1}^{K} (\bar{V}_1^* - \bar{V}_1^{\pi_k})(s_1^k)\\
    &\leq \sum_{k=1}^{\bar{K}} (V_1^k - V_1^{\pi_k})(s_1^k) + \sum_{k=\bar{K}+1}^{K} (V_1^k - \bar{V}_1^{\pi_k})(s_1^k).
\end{align}
For any fixed $(k,h)\in \{1,\cdots,\bar{K}\}\times [H]$, let $\delta_h^k := \left(V_h^k - V_h^{\pi_k}\right)(s_1^k)$ and $\phi_h^k := \left(V_h^k - V_h^*\right)(s_h^k)$, then we have
\begin{align}
    \delta_h^k &= \left(V_h^k - V_h^{\pi_k}\right)(s_1^k)\\
    &\leq \left(Q_h^k - Q_h^{\pi_k}\right)(s_h^k,a_h^k) \left(\text{by } V_h^k(s_h^k)\leq \max_{a'\in \mathcal{A}}Q_h^k(s_h^k,a')=Q_h^k (s_h^k,a_h^k)\right)\\
    &= \left(Q_h^k - Q_h^*\right)(s_h^k,a_h^k) + \left(Q_h^* - Q_h^{\pi_k}\right)(s_h^k,a_h^k)\\
    &\leq \alpha_t^0 H + \sum_{i=1}^t \alpha_t^i \phi_{h+1}^{k_i} + \beta_t + \left[\mathbb{P}_h \left(V_{h+1}^* - V_{h+1}^{\pi_k}\right)\right](s_h^k, a_h^k)\left(\text{by Lemma~\ref{lemma:boundQ}}\right)\\
    &=  \alpha_t^0 H + \sum_{i=1}^t \alpha_t^i \phi_{h+1}^{k_i} + \beta_t - \phi_{h+1}^k + \delta_{h+1}^k + \xi_{h+1}^k,
\end{align}
where $\beta_t=2\sum_{i=1}^t \alpha_t^ib_i\leq 4\frac{c}{p(\cdot|s,a;\theta_h)}\sqrt{H^3\iota/t}\leq 4\left(\frac{c}{p(\cdot|s,a;\theta_h)}+c\epsilon\right)\sqrt{H^3\iota/t}$ with $t=N_h(s,a)$, and $\xi_{h+1}^k:=\left[(\mathbb{P}_h - \hat{\mathbb{P}}_h^k)(V_{h+1}^* - V_{h+1}^k)\right](s_h^k, a_h^k)$ is a martingale difference sequence.

Similarly, by Lemma~\ref{lemma:boundQ}, for any fixed $(k,h)\in \{\bar{K}+1,\cdots,K\}\times [H]$, let $\bar{\delta}_h^k := \left(V_h^k - \bar{V}_h^{\pi_k}\right)(s_1^k)$ and $\bar{\phi}_h^k := \left(V_h^k - \bar{V}_h^*\right)(s_h^k)$, we have
\begin{align}
    \bar{\delta}_h^k =  \alpha_t^0 H + \sum_{i=1}^t \alpha_t^i \bar{\phi}_{h+1}^{k_i} + \beta_t - \bar{\phi}_{h+1}^k + \bar{\delta}_{h+1}^k + \bar{\xi}_{h+1}^k,
\end{align}
where $\bar{\xi}_{h+1}^k:=\left[(\bar{\mathbb{P}}_h - \hat{\mathbb{P}}_h^k)(\bar{V}_{h+1}^* - V_{h+1}^k)\right](s_h^k, a_h^k)$.

Hence, from Equation~\ref{eq:origin_regret}, the regret bound becomes
\begin{align}
    \text{Regret}(K) \leq \sum_{k=1}^{\bar{K}}\delta_h^k + \sum_{k=\bar{K}+1}^{K}\bar{\delta}_h^k.
\end{align}
Denoting by $n_h^k = N_h^k(s_h^k,a_h^k)$, then we have
\begin{align}
    \sum_{k=1}^{\bar{K}} \alpha_{n_h^k}^0 H + \sum_{k=\bar{K}+1}^{K} \alpha_{n_h^k}^0 H = \sum_{k=1}^K \alpha_{n_h^k}^0 H = \sum_{k=1}^K H\cdot \mathbb{I}[n_h^k=0] \leq |\mathcal{S}||\mathcal{A}|H.
\end{align}
Let $k_i(s_h^k,a_h^k)$ is the episode of which $(s_h^k,a_h^k)$ was taken at step $h$ for the $i-th$ time. For every $k'\in \{1,\cdots, \bar{K}\}$ and $k'\in \{\bar{K}+1,\cdots, K\}$, the corresponding term $\phi_{h+1}^{k'}$ and $\bar{\phi}_{h+1}^{k'}$ appears in the summand with $k>k'$ iff $(s_h^k,a_h^k)=(s_h^{k'},a_h^{k'})$. The first time it appears we have $n_h^k = n_h^{k'}+1$, the second time we have $n_h^k=n_h^{k'}+2$, etc. Hence, we obtain the following bound
\begin{align}
    &\sum_{k=1}^{\bar{K}}\sum_{i=1}^{n_h^k}\alpha_{n_h^k}^i \phi_{h+1}^{k_i(s_h^k,a_h^k)} + \sum_{k=\bar{K}+1}^{K}\sum_{i=1}^{n_h^k}\alpha_{n_h^k}^i \bar{\phi}_{h+1}^{k_i(s_h^k,a_h^k)} \leq \sum_{k'=1}^{\bar{K}} \phi_{h+1}^{k'}\sum_{t=n_h^{k'}+1}^\infty \alpha_t^{n_h^{k'}} + \sum_{k'=\bar{K}+1}^{K} \bar{\phi}_{h+1}^{k'}\sum_{t=n_h^{k'}+1}^\infty \alpha_t^{n_h^{k'}}\\
    &\leq \left(1+\frac{1}{H}\right) \left(\sum_{k=1}^{\bar{K}} \phi_{h+1}^{k} + \sum_{k=\bar{K}+1}^{K} \bar{\phi}_{h+1}^{k} \right) \left(\text{by $\sum_{t=i}^\infty \alpha_t^i = 1+\frac{1}{H}$ from Lemma~\ref{lemma:learningrate}}\right).
\end{align}
Therefore, the regret bound becomes
\small
\begin{align}
    \text{Regret}(K) &\leq \sum_{k=1}^{\bar{K}}\delta_h^k + \sum_{k=\bar{K}+1}^{K}\bar{\delta}_h^k\\
    &= \left(\sum_{k=1}^{\bar{K}} \alpha_t^0 H + \sum_{i=1}^t \alpha_t^i \phi_{h+1}^{k_i} + \beta_t - \phi_{h+1}^k + \delta_{h+1}^k + \xi_{h+1}^k\right)\nonumber\\
    &\quad+ \left(\sum_{k=\bar{K}+1}^{K}\alpha_t^0 H + \sum_{i=1}^t \alpha_t^i \bar{\phi}_{h+1}^{k_i} + \beta_t - \bar{\phi}_{h+1}^k + \bar{\delta}_{h+1}^k + \bar{\xi}_{h+1}^k\right)\\
    &\leq |\mathcal{S}||\mathcal{A}|H + \left(1+\frac{1}{H}\right) \left(\sum_{k=1}^{\bar{K}} \phi_{h+1}^{k} + \sum_{k=\bar{K}+1}^{K} \bar{\phi}_{h+1}^{k} \right) - \left(\sum_{k=1}^{\bar{K}}\phi_{h+1}^k + \sum_{k=\bar{K}+1}^{K}\bar{\phi}_{h+1}^k\right)\nonumber\\
    &\quad+ \left(\sum_{k=1}^{\bar{K}}\delta_{h+1}^k + \sum_{k=\bar{K}+1}^{K}\bar{\delta}_{h+1}^k\right) + \left(\sum_{k=1}^{\bar{K}} \beta_{n_h^k} + \sum_{k=\bar{K}+1}^K \beta_{n_h^k}\right) + \left(\sum_{k=1}^{\bar{K}} \xi_{h+1}^k + \sum_{k=\bar{K}+1}^K \bar{\xi}_{h+1}^k\right)\\
    &\leq |\mathcal{S}||\mathcal{A}|H + \left(1+\frac{1}{H}\right) \left(\sum_{k=1}^{\bar{K}} \phi_{h+1}^{k} + \sum_{k=\bar{K}+1}^{K} \bar{\phi}_{h+1}^{k} \right) + \left(\sum_{k=1}^{\bar{K}}\delta_{h+1}^k + \sum_{k=\bar{K}+1}^{K}\bar{\delta}_{h+1}^k\right)\nonumber\\ 
    &\quad+ \sum_{k=1}^K \beta_{n_h^k} + \left(\sum_{k=1}^{\bar{K}} \xi_{h+1}^k + \sum_{k=\bar{K}+1}^K \bar{\xi}_{h+1}^k\right).
\end{align}
\normalsize
Combining with the fact that $V^*\geq V^{\pi_k}$ and $\bar{V}^*\geq \bar{V}^{\pi_k}$, yielding $\phi_{h+1}^k \leq \delta_{h+1}^k$ and $\bar{\phi}_{h+1}^k \leq \bar{\delta}_{h+1}^k$, thus we get
\begin{align}
    \text{Regret}(K) \leq |\mathcal{S}||\mathcal{A}|H + \left(1+\frac{1}{H}\right) \left(\sum_{k=1}^{\bar{K}} \delta_{h+1}^{k} + \sum_{k=\bar{K}+1}^{K} \bar{\delta}_{h+1}^{k} \right) + \sum_{k=1}^K \beta_{n_h^k} + \left(\sum_{k=1}^{\bar{K}} \xi_{h+1}^k + \sum_{k=\bar{K}+1}^K \bar{\xi}_{h+1}^k\right).
\end{align}
Recursing the result for $h=1,2,\cdots,H$, and using the fact that $\delta_{H+1}^K=0$, we have
\begin{align}
    \text{Regret}(K) \leq \mathcal{O}\left(H^2|\mathcal{S}||\mathcal{A}| + \sum_{h=1}^H\sum_{k=1}^K \beta_{n_h^k} + \sum_{h=1}^H\left(\sum_{k=1}^{\bar{K}} \xi_{h+1}^k + \sum_{k=\bar{K}+1}^K \bar{\xi}_{h+1}^k\right)\right).
\end{align}
On the one hand, the r.v. $\xi_{h+1}^k:=\left[(\mathbb{P}_h - \hat{\mathbb{P}}_h^k)(V_{h+1}^* - V_{h+1}^k)\right](s_h^k, a_h^k)$ is a martingale difference sequence because if we define $\mathcal{F}_i$ be the $\sigma-$field generated by all the r.v. until episode $k_i$, step $h$, then, $\xi_{h+1}^k$ is a martingale difference sequence w.r.t the filtration $\{\mathcal{F}_i\}_{i\geq 0}$. Hence, by the Azuma-Hoeffding inequality, with probability $1-\delta$, we have 
\begin{align}
    \left|\sum_{h=1}^H\left(\sum_{k=1}^{\bar{K}} \xi_{h+1}^k + \sum_{k=\bar{K}+1}^K \bar{\xi}_{h+1}^k\right)\right|&=\left|\sum_{h=1}^H\left(\sum_{k=1}^{\bar{K}} \left[(\mathbb{P}_h - \hat{\mathbb{P}}_h^k)(V_{h+1}^* - V_{h+1}^k)\right](s_h^k, a_h^k)\right.\right.\nonumber\\
    &\quad + \left.\left.\sum_{k=\bar{K}+1}^K \left[(\bar{\mathbb{P}}_h - \hat{\mathbb{P}}_h^k)(\bar{V}_{h+1}^* - V_{h+1}^k)\right](s_h^k, a_h^k)\right)\right|\\
    &\leq \frac{c}{\min_{h\in [H]}p(\cdot|s,a;\theta_h)}H\sqrt{KH\iota} \leq \left(\frac{c}{\min_{h\in [H]}p(\cdot|s,a;\theta_h)}+c\epsilon\right)H\sqrt{KH\iota}.
\end{align}
On the other hand, by the pigeonhole principle,
\begin{align}
    \sum_{h=1}^H\sum_{k=1}^K \beta_{n_h^k} \leq  \sum_{h=1}^H \mathcal{O}(1+\epsilon) \sum_{k=1}^K \sqrt{\frac{H^3\iota}{n_h^k}}=\sum_{h=1}^H \mathcal{O}(1+\epsilon) \sum_{s,a}\sum_{n=1}^{N_h^K(s,a)}\sqrt{\frac{H^3\iota}{n}},
\end{align}
combining with the fact that $\sum_{s,a}N_h^K(s,a)=K$ and $\sum_{s,a}\sum_{n=1}^{N_h^K(s,a)}\sqrt{\frac{H^3\iota}{n}}$ is maximized when $N_h^K(s,a)=K/|\mathcal{S}||\mathcal{A}|$ for all $s,a$, we get
\begin{align}
    \sum_{h=1}^H\sum_{k=1}^K \beta_{n_h^k} \leq \mathcal{O}\left(\sqrt{H^5(1+\epsilon)^2|\mathcal{S}||\mathcal{A}|K\iota}\right).
\end{align}
Note that when $KH\geq \sqrt{H^5(1+\epsilon)^2|\mathcal{S}||\mathcal{A}|T\iota}$, by infinitely nested radical, we have $\sqrt{H^5(1+\epsilon)^2\mathcal{S}||\mathcal{A}|K\iota}\geq H^2|\mathcal{S}||\mathcal{A}|$. And, when $KH\leq \sqrt{H^5(1+\epsilon)^2|\mathcal{S}||\mathcal{A}|K\iota}$, by the worst case of regret is $KH$, we have $\sum_{k=1}^{\bar{K}}\delta_h^k + \sum_{k=\bar{K}+1}^{K}\bar{\delta}_h^k\leq KH\leq \sqrt{H^5(1+\epsilon)^2|\mathcal{S}||\mathcal{A}|K\iota}$. Therefore, by $\iota = \log(|\mathcal{S}| |\mathcal{A}|KH/\delta)$, with probability at least $1-\delta$, we obtain
\begin{align}
    \text{Regret}(K) \leq \sum_{k=1}^{\bar{K}}\delta_h^k + \sum_{k=\bar{K}+1}^{K}\bar{\delta}_h^k \leq \mathcal{O}\left(\sqrt{H^5(1+\epsilon)^2|\mathcal{S}||\mathcal{A}|K \log(|\mathcal{S}| |\mathcal{A}|KH/\delta)}\right)
\end{align}
of Theorem~\ref{thm:episode_bound}.
\end{proof}

\subsection{Proof of Theorem~\ref{thm:infiteMDP_bound}}\label{apd:thm:infiteMDP_bound}
\begin{proof}
    Similar to the proof in Theorem~\ref{thm:episode_bound}, this proof is based on the bound on learning error of $Q$-function in the infinite-horizon discounted MDP setting. Firstly, from Lemma~\ref{lemma:boundQ2}, we can see that the different between the bound of $(\hat{Q}^t-Q^*)(s,a)$ and $(\hat{Q}^t-\bar{Q}^*)(s,a)$ is only in $V^*$ and $\bar{V}^*$. Therefore, we can borrow Theorem 7.3 of~\citet{yang2021qlearning}, i.e., with probability at least $1-\delta$, where $\delta = 1/T$, for every $n\in [\lceil \log_2(1/\Delta_{\min}(1-\gamma)) \rceil]$, we have
    \begin{align}\label{eq:boundC}
        C^{(n)}&:=\left|\left\{t\in \mathbb{N}_+: \left(\hat{Q}^t-Q^*\right)(s_t,a_t)\in \left[2^{n-1} \Delta_{\min}, 2^n \Delta_{\min}\right]\right\}\right|\\ 
        &\leq \mathcal{O}\left(\frac{|\mathcal{S}||\mathcal{A}| (1+\epsilon)}{4^n(1-\gamma)^5\Delta_{\min}^2}\ln\left(\frac{|\mathcal{S}||\mathcal{A}|T}{(1-\gamma)\Delta_{\min}}\right)\right),
    \end{align}
    and for every $n\in [\lceil \log_2(1/\bar{\Delta}_{\min}(1-\gamma)) \rceil]$,
    \begin{align}\label{eq:boundC2}
        \bar{C}^{(n)}&:=\left|\left\{t\in \mathbb{N}_+: \left(\hat{Q}^t-\bar{Q}^*\right)(s_t,a_t)\in \left[2^{n-1} \bar{\Delta}_{\min}, 2^n \bar{\Delta}_{\min}\right]\right\}\right|\\ 
        &\leq \mathcal{O}\left(\frac{|\mathcal{S}||\mathcal{A}|(1+\epsilon)}{4^n(1-\gamma)^5\bar{\Delta}_{\min}^2}\ln\left(\frac{|\mathcal{S}||\mathcal{A}|T}{(1-\gamma)\bar{\Delta}_{\min}}\right)\right).
    \end{align}

    By the definition of the sub-optimality gap, we have
    \begin{align}
        \text{Regret}(T) &= \sum_{t=1}^{\bar{T}} \left(V^* - V^{\pi_t}\right)(s_t) + \sum_{t=\bar{T}+1}^{T} \left(\bar{V}^* - \bar{V}^{\pi_t}\right)(s_t)\\
        &=\sum_{t=1}^{\bar{T}} \mathbb{E}\left[\sum_{h=0}^\infty \gamma^h \Delta(s_{t+h},a_{t+h})\Big| a_{t+h}=\pi_{t+h}(s_{t+h})\right]\nonumber\\
        &\quad + \sum_{t=\bar{T}+1}^{T} \mathbb{E}\left[\sum_{h=0}^\infty \gamma^h \bar{\Delta}(s_{t+h},a_{t+h})\Big| a_{t+h}=\pi_{t+h}(s_{t+h})\right]\\
        &= \sum_{t=1}^{\bar{T}}\mathbb{E}\left[\sum_{h=0}^\infty \gamma^h \Delta(s_{t+h},a_{t+h})\right] + \sum_{t=\bar{T}+1}^{T}\mathbb{E}\left[\sum_{h=0}^\infty \gamma^h \bar{\Delta(}s_{t+h},a_{t+h})\right]\\
        &= \sum_{t=1}^{\bar{T}}\sum_{h'=t}^\infty \gamma^{h'-t} \Delta(s_{h'},a_{h'}) + \sum_{t=\bar{T}+1}^{T}\sum_{h'=t}^\infty \gamma^{h'-t} \bar{\Delta}(s_{h'},a_{h'}).
    \end{align}
    
    For a fixed $s_t$, for every infinite-length trajectory $traj$, the trajectories inside the event in which all the learning errors of the value function is both bounded below (by zero) and bounded above, we can apply the concentration bound with Azuma–Hoeffding to the sub-optimality gap~\citep{yang2021qlearning}, whereas for trajectories outside of this event, the sub-optimality gaps never exceed the time horizon. Hence, by the Azuma–Hoeffding inequality, with probability at least $1-\delta$, we have
    \begin{align}
        \text{Regret}(T) &\leq \sum_{t=1}^{\bar{T}}\sum_{traj}p(traj)\left[\sum_{h'=t}^{\infty}\gamma^{h'-t} \Delta(s_{h'},a_{h'})\right] + \sum_{t=\bar{T}+1}^{T}\sum_{traj}p(traj)\left[\sum_{h'=t}^{\infty}\gamma^{h'-t} \bar{\Delta}(s_{h'},a_{h'})\right]\\
        &= \sum_{t=1}^{\bar{T}}\sum_{h'=t}^{\infty}\gamma^{h'-t} \Delta(s_{h'},a_{h'}|traj) + \sum_{t=\bar{T}+1}^{T}\sum_{h'=t}^{\infty}\gamma^{h'-t} \bar{\Delta}(s_{h'},a_{h'}|traj)\\ 
        &= \sum_{h=1}^\infty \Delta(s_h,a_h)\sum_{t=1}^{\min\{\bar{T},h\}}\gamma^t + \sum_{h=1}^\infty \bar{\Delta}(s_h,a_h)\sum_{t=\bar{T}+1}^{\min\{T,h\}}\gamma^t\\
        &\leq 2\max\left\{\frac{1}{1-\gamma} \sum_{h=1}^\infty \Delta(s_h,a_h|traj),\frac{1}{1-\gamma} \sum_{h=1}^\infty \bar{\Delta}(s_h,a_h|traj)\right\}\\
        &\quad\left(\text{By the optimism of estimated $Q$-values}\right)\nonumber.
    \end{align}
    Since we can add an outer summation over sub-intervals $n\in [N]$ and bound each of them by their maximum value times the number of steps inside, we get
    \begin{align}
        \text{Regret}(T) 
        &\leq 2\max\left\{\frac{1}{1-\gamma} \sum_{n=1}^N 2^n\Delta_{\min}C^{(n)}, \frac{1}{1-\gamma} \sum_{n=1}^N 2^n\bar{\Delta}_{\min}\bar{C}^{(n)}\right\}.
    \end{align}
    Using results in Equation~\ref{eq:boundC} and Equation~\ref{eq:boundC2}, we obtain
    \begin{align}
        \text{Regret}(T) &\leq \max\left\{\mathcal{O}\left(\frac{|\mathcal{S}| |\mathcal{A}|(1+\epsilon)}{(1-\gamma)^6\Delta_{\min}} \log\left(\frac{|\mathcal{S}| |\mathcal{A}|T}{(1-\gamma)\Delta_{\min}}\right)\right), \mathcal{O}\left(\frac{|\mathcal{S}| |\mathcal{A}|(1+\epsilon)}{(1-\gamma)^6\bar{\Delta}_{\min}} \log\left(\frac{|\mathcal{S}| |\mathcal{A}|T}{(1-\gamma)\bar{\Delta}_{\min}}\right)\right)\right\}\\
        &= \mathcal{O}\left(\frac{|\mathcal{S}| |\mathcal{A}|(1+\epsilon)}{(1-\gamma)^6\min\{\Delta_{\min}, \bar{\Delta}_{\min}\}} \log\left(\frac{|\mathcal{S}| |\mathcal{A}|T}{(1-\gamma)\min\{\Delta_{\min}, \bar{\Delta}_{\min}\}}\right)\right).
    \end{align}
    of Theorem~\ref{thm:infiteMDP_bound}.
\end{proof}

\subsection{Proof of Lemma~\ref{lemma:boundQ}}\label{proof:lemma:boundQ}
\begin{proof}
    For any $(s,a,h)\in \mathcal{S} \times \mathcal{A} \times [H]$ and episode $k\in [K]$, from the Bellman optimality equation, $Q_h^*(s,a)=(r_h+\mathbb{P}_h V_{h+1}^*)(s,a)$ and $\bar{Q}_h^*(s,a)=(r_h+\bar{\mathbb{P}}_h \bar{V}_{h+1}^*)(s,a)$, since $\left[\hat{\mathbb{P}}_h^{k_i} V_{h+1}\right](s,a) = V_{h+1}(s_{h+1}^{k_i})$, we have $\left[\hat{\mathbb{P}}_h^{k_i} \bar{V}_{h+1}\right](s,a) = \bar{V}_{h+1}(s_{h+1}^{k_i})$, thus
    \begin{align}
        &Q_h^*(s,a) = \alpha_t^0 Q_h^*(s,a) + \sum_{i=1}^t \alpha_t^i \left[r_h(s,a) + (\mathbb{P}_h - \hat{\mathbb{P}}_h^{k_i})V_{h+1}^*(s,a) + V_{h+1}^*(s_{h+1}^{k_i})\right],\\
        &\bar{Q}_h^*(s,a) = \alpha_t^0 \bar{Q}_h^*(s,a) + \sum_{i=1}^t \alpha_t^i \left[r_h(s,a) + (\bar{\mathbb{P}}_h - \hat{\mathbb{P}}_h^{k_i})\bar{V}_{h+1}^*(s,a) + \bar{V}_{h+1}^*(s_{h+1}^{k_i})\right].
    \end{align}
    Subtracting to $Q_{h}^k(s,a)$ in Equation~\ref{eq:empiricalQ}, we obtain 
    \begin{align}\label{eq:recursionQ}
        &(Q_{h}^k - Q_h^*)(s,a) = \alpha_t^0(H-Q_h^*(s,a)) + \sum_{i=1}^t \alpha_t^i \left[(V_{h+1}^{k_i} - V_{h+1}^*)(s_{h+1}^{k_i}) + \left[(\hat{\mathbb{P}}_h^{k_i} - \mathbb{P}_h)V_{h+1}^*\right](s,a) + b_i\right],\\
        &(Q_{h}^k - \bar{Q}_h^*)(s,a) = \alpha_t^0(H-\bar{Q}_h^*(s,a)) + \sum_{i=1}^t \alpha_t^i \left[(V_{h+1}^{k_i} - \bar{V}_{h+1}^*)(s_{h+1}^{k_i}) + \left[(\hat{\mathbb{P}}_h^{k_i} - \bar{\mathbb{P}}_h)\bar{V}_{h+1}^*\right](s,a) + b_i\right].
    \end{align}
If $t=N_h^k(s,a)\in \{1,\cdots, \bar{K}-1\}$, the transition operator at step $h$ is $\mathbb{P}_h$, then $p(\cdot|s,a;\theta_h)=\mathbb{P}_h/\mathbb{P}_h$, and for all $(s,a,h,k)\in \mathcal{S} \times \mathcal{A} \times [H]\times \{1,\cdots, \bar{K}-1\}$, from Lemma~4.3~\citep{jin2018isQ}, with probability at least $1-\delta$, we have
\begin{align}\label{eq:halfVbound}
    \left|\sum_{i=1}^t \alpha_t^i \left[(\hat{\mathbb{P}}_h^{k_i}-\mathbb{P}_h)V_{h+1}^*\right](s,a)\right|\leq c\frac{\mathbb{P}_h}{\mathbb{P}_h}\sqrt{\frac{H^3\iota}{t}}=\frac{c}{p(\cdot|s,a;\theta_h)}\sqrt{\frac{H^3\iota}{t}}.
\end{align}

Consider the transition operator at step $h$ change from $\mathbb{P}_h$ to $\bar{\mathbb{P}}_h$ at episode $\bar{K}$, then $p(\cdot|s,a;\theta_h)=\mathbb{P}_h/\mathbb{P}_h$ before $\bar{K}$, $p(\cdot|s,a;\theta_h)=\mathbb{P}_h/\bar{\mathbb{P}}_h$ at $\bar{K}$, and for $t=\bar{K}$, $\forall (s,a,h,k)\in \mathcal{S} \times \mathcal{A} \times [H]\times \{\bar{K}\}$, we have
\begin{align}
    \left|\sum_{i=1}^t \alpha_t^i \left[(\hat{\mathbb{P}}_h^{k_i}-\bar{\mathbb{P}}_h)\bar{V}_{h+1}^*\right](s,a)\right| = \left|\sum_{i=1}^t \alpha_t^i \left[(\hat{\mathbb{P}}_h^{k_i}-\mathbb{P}_h)V_{h+1}^*\right](s,a)\right| \frac{\bar{\mathbb{P}}_h}{\mathbb{P}_h}.
\end{align}
Hence, using the result in Equation~\ref{eq:halfVbound}, we get
\begin{align}\label{eq:halfVbound2}
    \left|\sum_{i=1}^t \alpha_t^i \left[(\hat{\mathbb{P}}_h^{k_i}-\bar{\mathbb{P}}_h)\bar{V}_{h+1}^*\right](s,a)\right| &\leq c\frac{\mathbb{P}_h}{\mathbb{P}_h}\frac{\bar{\mathbb{P}}_h}{\mathbb{P}_h}\sqrt{\frac{H^3\iota}{t}}=c\frac{\bar{\mathbb{P}}_h}{\mathbb{P}_h}\sqrt{\frac{H^3\iota}{t}}=\frac{c}{p(\cdot|s,a;\theta_h)}\sqrt{\frac{H^3\iota}{t}}.
\end{align}

And, when the transition operator at step $h$ is $\bar{\mathbb{P}}_h$ after $\bar{K}$, then $p(\cdot|s,a;\theta_h)=\bar{\mathbb{P}}_h/\bar{\mathbb{P}}_h$, and for any $t\in \{\bar{K}+1,\cdots,K\}$, $\forall (s,a,h,k)\in \mathcal{S} \times \mathcal{A} \times [H]\times \{\bar{K}+1,\cdots,K\}$, with probability at least $1-\delta$, we have
\begin{align}\label{eq:Vbarbound}
    \left|\sum_{i=\bar{K}+1}^t \alpha_t^i \left[(\hat{\mathbb{P}}_h^{k_i}-\bar{\mathbb{P}}_h)\bar{V}_{h+1}^*\right](s,a)\right|\leq c\frac{\bar{\mathbb{P}}_h}{\bar{\mathbb{P}}_h}\sqrt{\frac{H^3\iota}{t}} = \frac{c}{p(\cdot|s,a;\theta_h)}\sqrt{\frac{H^3\iota}{t}}.
\end{align}

Combining the result in Equation~\ref{eq:halfVbound}, Equation~\ref{eq:halfVbound2}, and Equation~\ref{eq:Vbarbound}, we obtain for any $t\in [K]$, $\forall (s,a,h,k)\in \mathcal{S} \times \mathcal{A} \times [H]\times [K]$, with probability at least $1-\delta$, it holds that
\begin{align}\label{eq:boundPV}
    \left|\sum_{i=1}^t \alpha_t^i \left[(\hat{\mathbb{P}}_h^{k_i}-\mathbb{P}_h)V_{h+1}^*\right](s,a)\right|\leq \frac{c}{p(\cdot|s,a;\theta_h)}\sqrt{\frac{H^3\iota}{t}}, \text{ where } t=N_h^k(s,a).
\end{align}
Note that if we choose $b_t = \frac{c}{p(\cdot|s,a;\theta_h)}\sqrt{H^3\iota/t}$, i.e., $\beta_t=2\sum_{i=1}^t \alpha_t^ib_i\leq 4\frac{c}{p(\cdot|s,a;\theta_h)}\sqrt{H^3\iota/t}$, by Lemma~\ref{lemma:learningrate}, $\beta_t/2=\sum_{i=1}^t\alpha_t^ib_i\in \left[\frac{c}{p(\cdot|s,a;\theta_h)}\sqrt{\frac{H^3\iota}{t}}, \frac{2c}{p(\cdot|s,a;\theta_h)}\sqrt{\frac{H^3\iota}{t}}\right]$, combining with results in Equation~\ref{eq:recursionQ} and Equation~\ref{eq:boundPV}, we obtain
\begin{align}\label{eq:proof:episode_bound:RHS}
    &(Q_h^k-Q_h^*)(s,a) \leq \alpha_t^0 H +\sum_{i=1}^t \alpha_t^i \left(V_{h+1}^{k_i} - V_{h+1}^*\right)(s_{h+1}^{k_i}) + \beta_t,\\
    &(Q_h^k-\bar{Q}_h^*)(s,a) \leq \alpha_t^0 H +\sum_{i=1}^t \alpha_t^i \left(V_{h+1}^{k_i} - \bar{V}_{h+1}^*\right)(s_{h+1}^{k_i}) + \beta_t.
\end{align}
On the other hand, applying induction on $h=H,H-1,\cdots,1$ to results in Equation~\ref{eq:recursionQ} and result in Equation~\ref{eq:boundPV}, we obtain 
\begin{align}\label{eq:proof:episode_bound:LHS}
    0\leq (Q_h^k-Q_h^*)(s,a), \quad 0\leq (Q_h^k-\bar{Q}_h^*)(s,a).
\end{align}
Combining results in Equation~\ref{eq:proof:episode_bound:RHS} and Equation~\ref{eq:proof:episode_bound:LHS}, we obtain the results of Lemma~\ref{lemma:boundQ}.
\end{proof}

\subsection{Proof of Lemma~\ref{lemma:boundQ2}}\label{proof:lemma:boundQ2}
\begin{proof}
    For any $(s,a)\in \mathcal{S} \times \mathcal{A}$ and time step $t\in [T]$, from the Bellman optimality equation, we have
    \begin{align}
        &Q^*(s,a) = r(s,a) + \gamma \mathbb{P}V^*(s,a) = \alpha_t^0 Q^*(s,a) + \sum_{t=1}^t \alpha_t^i \left[r(s,a) + \gamma \mathbb{P}V^*(s,a)\right],\\
        &\bar{Q}^*(s,a) = r(s,a) + \gamma \bar{\mathbb{P}}\bar{V}^*(s,a) = \alpha_t^0 \bar{Q}^*(s,a) + \sum_{t=1}^t \alpha_t^i \left[r(s,a) + \gamma \bar{\mathbb{P}}\bar{V}^*(s,a)\right].
    \end{align}
    Subtracting to $\hat{Q}^t(s,a)$ in Equation~\ref{eq:empiricalQ2}, we obtain
    \begin{align}\label{eq:recursionQ2}
        \left(\hat{Q}^t - Q^*\right)(s,a) &= \alpha_t^0 \left(\frac{1}{1-\gamma} - Q^*(s,a)\right)\nonumber\\
        &\quad+ \sum_{i=1}^t \alpha_t^i \left[ \gamma(V_{t_1}-V^*)(s_{t_i+1}) + \gamma\left(V^*(s_{t_i+1}) - \mathbb{P}V^*(s,a)\right) + b_i \right],\\
        \left(\hat{Q}^t - \bar{Q}^*\right)(s,a) &= \alpha_t^0 \left(\frac{1}{1-\gamma} - \bar{Q}^*(s,a)\right)\nonumber\\
        &\quad+ \sum_{i=1}^t \alpha_t^i \left[ \gamma(V_{t_1}-\bar{V}^*)(s_{t_i+1}) + \gamma\left(\bar{V}^*(s_{t_i+1}) - \Bar{\mathbb{P}}\bar{V}^*(s,a)\right) + b_i \right],
    \end{align}
    If $t=N_p(s,a)\in \{1,\cdots, \bar{T}-1\}$, $t_i=\tau(s,a,i)$, the transition operator at step $t$ is $\mathbb{P}$, then $p(\cdot|s,a;\theta)=\mathbb{P}/\mathbb{P}$, and for all $(s,a,t)\in \mathcal{S} \times \mathcal{A} \times \{1,\cdots, \bar{T}-1\}$, from Lemma~4~\citep{wang2020Qlearning}, with probability at least $1-\delta$, we have
    \begin{align}\label{eq:halfVbound3}
        \gamma\left|\sum_{i=1}^t \left(\alpha_k^i \mathbb{I}[t_i<\infty] \left(\mathbb{P} - \hat{\mathbb{P}}_{t_i} \right)V^*(s,a)\right) \right|\leq \frac{3c_2}{1-\gamma}\frac{\mathbb{P}}{\mathbb{P}}\sqrt{\frac{H\iota(t)}{t}}=\frac{3c_2}{(1-\gamma)p(\cdot|s,a;\theta)}\sqrt{\frac{H\iota(t)}{t}}.
    \end{align}
     Consider the transition operator at step $t$ change from $\mathbb{P}$ to $\bar{\mathbb{P}}$ at time step $\bar{T}$, then $p(\cdot|s,a;\theta)=\mathbb{P}/\mathbb{P}$ before $\bar{T}$, $p(\cdot|s,a;\theta)=\mathbb{P}/\bar{\mathbb{P}}$ at $\bar{T}$, and for any $t = \bar{T}$, $\forall (s,a,t)\in \mathcal{S} \times \mathcal{A} \times \{\bar{T}\}$, we have
    \begin{align}
        \gamma\left|\sum_{i=1}^t \left(\alpha_k^i \mathbb{I}[t_i<\infty] \left(\bar{\mathbb{P}} - \hat{\mathbb{P}}_{t_i}\right) \bar{V}^*(s,a)\right) \right| &= \gamma\left|\sum_{i=1}^t  \left(\alpha_k^i \mathbb{I}[t_i<\infty] \left(\mathbb{P} - \hat{\mathbb{P}}_{t_i} \right)V^*(s,a)\right)\right| \frac{\bar{\mathbb{P}}}{\mathbb{P}}.
    \end{align}
    Hence, using the result in Equation~\ref{eq:halfVbound3}, we get
    \begin{align}\label{eq:halfVbound4}
        \gamma\left|\sum_{i=1}^t \left(\alpha_k^i \mathbb{I}[t_i<\infty] \left(\bar{\mathbb{P}} - \hat{\mathbb{P}}_{t_i}\right) \bar{V}^*(s,a)\right) \right| &\leq \frac{3c_2}{(1-\gamma)}\frac{\mathbb{P}}{\mathbb{P}}\frac{\bar{\mathbb{P}}}{\mathbb{P}}\sqrt{\frac{H\iota(t)}{t}} =\frac{3c_2}{(1-\gamma)}\frac{\bar{\mathbb{P}}}{\mathbb{P}}\sqrt{\frac{H\iota(t)}{t}}\\
        &=\frac{3c_2}{(1-\gamma)p(\cdot|s,a;\theta)}\sqrt{\frac{H\iota(t)}{t}}.
    \end{align}
     
    And, when the transition operator at step $t$ is $\bar{\mathbb{P}}$ after $\bar{T}$, then $p(\cdot|s,a;\theta)=\bar{\mathbb{P}}/\bar{\mathbb{P}}$, and for any $t\in \{\bar{T}+1,\cdots,T\}$, $\forall (s,a,t)\in \mathcal{S} \times \mathcal{A} \times \{\bar{T}+1,\cdots,T\}$, with probability at least $1-\delta$, we have
    \begin{align}\label{eq:Vbarbound2}
        \gamma\left|\sum_{i=\bar{T}+1}^t \left(\alpha_k^i \mathbb{I}[t_i<\infty] \left(\bar{\mathbb{P}} - \hat{\mathbb{P}}_{t_i}\right) \bar{V}^*(s,a)\right) \right|\leq \frac{3c_2}{1-\gamma}\frac{\bar{\mathbb{P}}}{\bar{\mathbb{P}}}\sqrt{\frac{H\iota(t)}{t}}=\frac{3c_2}{(1-\gamma)p(\cdot|s,a;\theta)}\sqrt{\frac{H\iota(t)}{t}}.
    \end{align}
    
    Combining the result in Equation~\ref{eq:halfVbound3}, Equation~\ref{eq:halfVbound4}, and Equation~\ref{eq:Vbarbound2}, we obtain for any $t\in [K]$, $\forall (s,a,t)\in \mathcal{S} \times \mathcal{A} \times[T]$, with probability at least $1-\delta$, it holds that
    \begin{align}\label{eq:boundPV2}
         \gamma\left|\sum_{i=1}^t \left(\alpha_k^i \mathbb{I}[t_i<\infty] \left(\mathbb{P} - \hat{\mathbb{P}}_{t_i} \right)V^*(s,a)\right) \right|\leq  \frac{3c_2}{(1-\gamma)p(\cdot|s,a;\theta)}\sqrt{\frac{H\iota(t)}{t}}, \text{ where } t=N_p(s,a).
    \end{align}
    Note that if we choose $b_t = \frac{c_2}{(1-\gamma)p(\cdot|s,a;\theta)}\sqrt{\frac{H\iota(t)}{t}}$, i.e., $\beta_t=2\sum_{i=1}^t \alpha_t^ib_i\leq 4\frac{c_2}{(1-\gamma)p(\cdot|s,a;\theta)}\sqrt{\frac{H\iota(t)}{t}}$, by Lemma~\ref{lemma:learningrate}, $\beta_t/2=\sum_{i=1}^t\alpha_t^ib_i\in \left[ \frac{c_2}{(1-\gamma)p(\cdot|s,a;\theta)}\sqrt{\frac{H\iota(t)}{t}},  \frac{2c_2}{(1-\gamma)p(\cdot|s,a;\theta)}\sqrt{\frac{H\iota(t)}{t}}\right]$, combining with results in Equation~\ref{eq:recursionQ2} and Equation~\ref{eq:boundPV2}, we obtain
    \begin{align}\label{eq:proof:episode_bound:RHS2}
        (\hat{Q}^t-Q^*)(s,a) &\leq \frac{\alpha_t^0}{1-\gamma}+\gamma\left|\sum_{i=1}^t \left(\alpha_k^i \mathbb{I}[t_i<\infty] \left(\mathbb{P} - \hat{\mathbb{P}}_{t_i} \right)V^*(s,a)\right) \right|\nonumber\\
        &\quad+ \sum_{i=1}^t \alpha_t^i \left[\gamma(\hat{V}_{t_1}-V^*)(s_{t_i+1})+b_i\right]\\
        &\leq \frac{\alpha_t^0}{1-\gamma} + \sum_{i=1}^t \gamma \alpha_t^i \left(\hat{V}_{t_i} - V^*\right)(s_{t_i+1}) + \beta_t,
    \end{align}
    and
    \begin{align}
       (\hat{Q}^t-\bar{Q}^*)(s,a) &\leq \frac{\alpha_t^0}{1-\gamma}+\gamma\left|\sum_{i=1}^t \left(\alpha_k^i \mathbb{I}[t_i<\infty] \left(\bar{\mathbb{P}} - \hat{\mathbb{P}}_{t_i} \right) \bar{V}^*(s,a)\right) \right|\nonumber\\ 
       &\quad+ \sum_{i=1}^t \alpha_t^i \left[\gamma(\hat{V}_{t_1}-\bar{V}^*)(s_{t_i+1})+b_i\right]\\
        &\leq \frac{\alpha_t^0}{1-\gamma} + \sum_{i=1}^t \gamma \alpha_t^i \left(\hat{V}_{t_i} - \bar{V}^*\right)(s_{t_i+1}) + \beta_t.
    \end{align}
    On the other hand, applying induction on $t = T, T-1, \cdots, 1$ to results in Equation~\ref{eq:recursionQ2} and result in Equation~\ref{eq:boundPV2}, we obtain 
    \begin{align}\label{eq:proof:episode_bound:LHS2}
        0\leq (\hat{Q}^t-Q^*)(s,a), \quad 0\leq (\hat{Q}^t-\bar{Q}^*)(s,a).
    \end{align}
    Combining results in Equation~\ref{eq:proof:episode_bound:RHS2} and Equation~\ref{eq:proof:episode_bound:LHS2}, we obtain the results of Lemma~\ref{lemma:boundQ2}.
\end{proof}

\section{Experimental details}\label{apd:exp}
\begin{figure}[ht!]
    \centering
    \hspace*{-0.1in}
    \setlength{\tabcolsep}{0.1pt}
    \begin{tabular}{ccc}
    \includegraphics[width=0.35\linewidth]{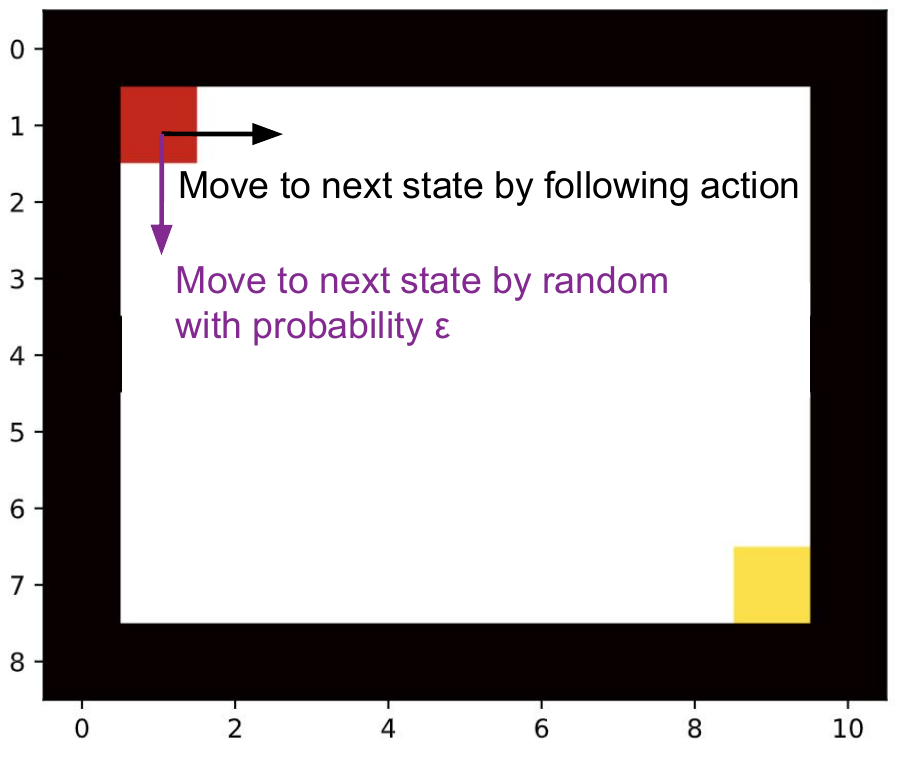}&
    \includegraphics[width=0.3\linewidth]{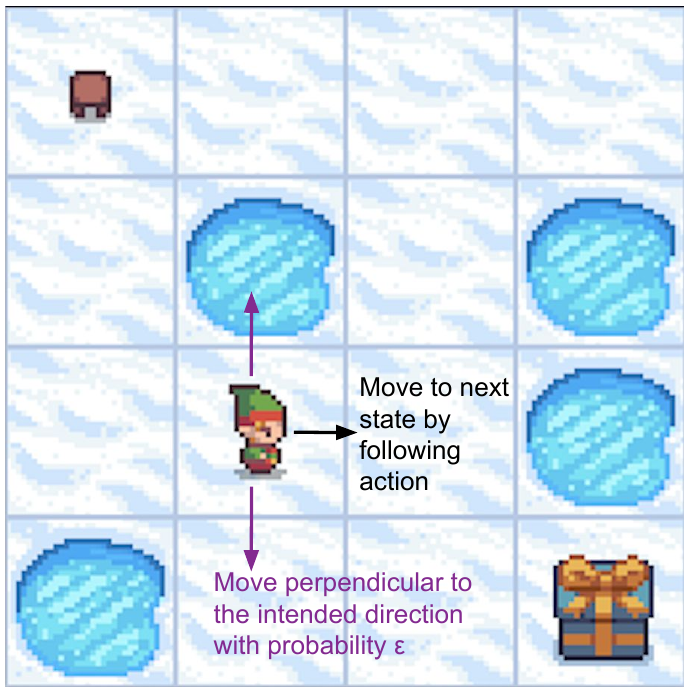}&
    \includegraphics[width=0.35\linewidth]{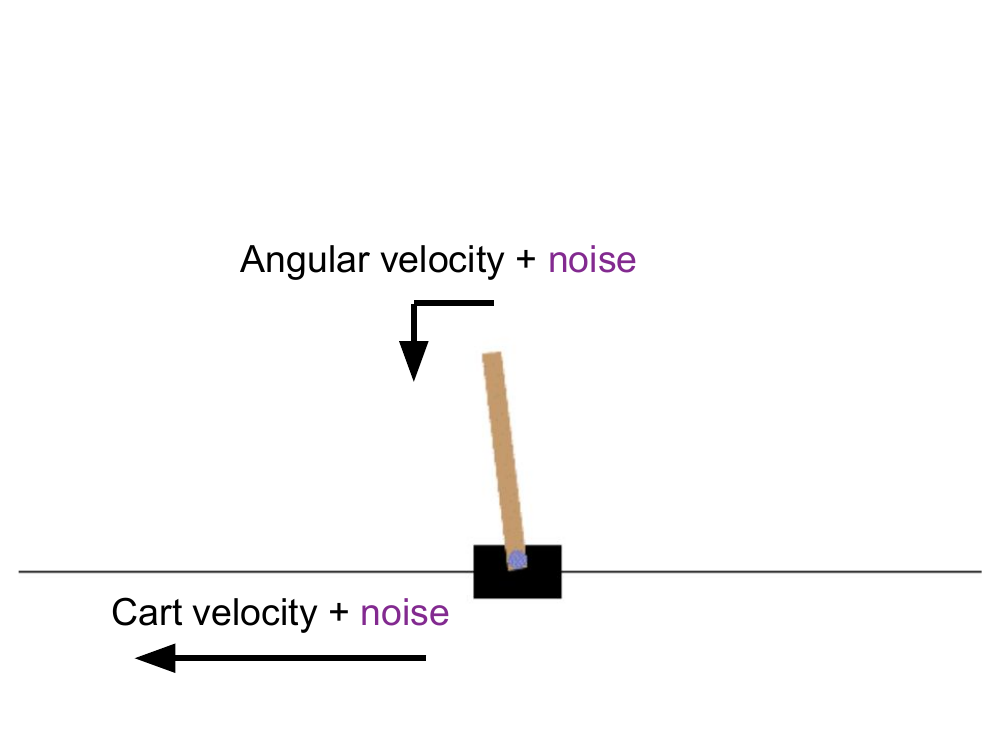}\\
    (a) & (b) & (c)
    \end{tabular}
    \caption{(a) The GridWorld task: the starting state is shown in red, the rewarding state is shown in yellow, and the transitions are noisy with $\epsilon$; (b) The Frozen-Lake task: the starting state is the chair at the top-left corner, the rewarding state is the box at the bottom-right corner, the goal is crossing a frozen lake from start to the rewarding state without falling into any holes by walking over the frozen lake, and the transitions are noisy with a slippery level $\epsilon$; (c) The CartPole task: the pendulum is placed upright on the cart, the goal is to balance the pole, and the transition noise $\mathcal{N}(0,0.15)$ is added to the velocity state.}
    \label{fig:dataset}
\end{figure}

\subsection{Experimental settings}\label{apd:exp_settings}
\subsubsection{GridWorld}
Fig.~\ref{fig:dataset}~(a) shows our $[10]\times[5]$ GridWorld environment with $50$ states and $4$ actions (left, right, up, down). When the agent uses its policy to take action, it will move in the corresponding direction following the action with probability $1-\epsilon$, and move to a neighbor state at random with probability $\epsilon$. The starting position is $(1, 1)$. The reward is equal to $1$ in state $(10, 5)$ and is zero elsewhere. We set the planning horizon $H=100$ and the number of episodes $K=50000$. From the first episode to $\Bar{K}=25000$, the noise is $\epsilon=0.01$. After that, the transition function is shifted by changing $\epsilon=0.2$.

\textbf{State}: The state $s_t=(i,j) \in [10]\times [5]$ at step $t$ consists of the agent's position on the grid, where $i$ is the row and $j$ is the column index. \textbf{Action}: Action $a_t\in \{left,right,up,down\}$ at time $t$ is the direction in which the agent will try to reach. \textbf{State Transition Function}: From position state $s_t$, the next position of agent $s_{t+1}$ will follow action $a_t$ with probability $1-\epsilon$, and move to a random neighbor position of $s_t$ with probability $\epsilon$. \textbf{Reward Function}: The reward $r_t$ equals $1$ when the agent reaches the final state $(10,5)$, and equals $0$ elsewhere.

\textbf{Baselines}: We compare our method with: (1) Q-learning UCB (QUCB)~\citep{jin2018isQ}; (2) Q-learning UCB, but additionally incorporates a momentum term that is built from the value functions for each state-action pair (UCBMQ)~\citep{menard2021UCBmomentum}; (3) a model-based RL baseline which directly adds a UCB bonus to the Q-values (UCBVI)~\citet{azar2017minimax}.

\subsubsection{Frozen-Lake}
Fig.~\ref{fig:dataset}~(c) shows our $[4]\times[4]$ FrozenLake environment with $16$ states and $4$ actions (left, right, up, down). This is a more challenging setting than GridWorld, as it involves crossing a frozen lake from the start to the goal without falling into any holes by walking across the frozen lake. When the agent uses its policy to take action, it will move in the corresponding direction following the action with probability $1-\epsilon$, and move in either perpendicular direction with equal probability (i.e., $\epsilon/2$) in both directions. The starting position is $(1, 1)$. The reward is equal to $1$ in state $(4, 4)$ and is zero elsewhere. The episode ends if the agent moves to the hole or reaches the goal. We set the planning horizon $H=500$ and the number of episodes $K=60000$. From the first episode to $\Bar{K}=20000$, the noise is $\epsilon=0$. Then, from episode $20000$-th to $\Bar{K}=40000$, the noise $\epsilon=1/2$. After that, the transition function is shifted by changing $\epsilon=2/3$.

\textbf{State}: The state $s_t=(i,j) \in [8]\times [8]$ at step $t$ consists of the agent's position on the grid, where $i$ is the row and $j$ is the column index. \textbf{Action}: Action $a_t\in \{left,right,up,down\}$ at time $t$ is the direction in which the agent will try to reach. \textbf{State Transition Function}: From position state $s_t$, the next position of the agent $s_{t+1}$ will follow action $a_t$ with probability $1-\epsilon$, and move in either perpendicular direction with equal probability (i.e., $\epsilon/2$) in both directions. \textbf{Reward Function}: The reward $r_t$ equals $1$ when the agent reaches the final state $(8,8)$, and equals $0$ elsewhere.

\textbf{Baselines}: Similarly to the GridWorld task, we compare our method with the following tabular UCB-based RL baselines: (1) QUCB~\citep{jin2018isQ}; (2) UCBMQ~\citep{menard2021UCBmomentum}; (3) UCBVI~\citet{azar2017minimax}.

\subsubsection{CartPole}
Fig.~\ref{fig:dataset}~(c) shows this task with 4 continuous state features (cart position, cart velocity, pole angle, pole angular velocity), and 2 actions (left, right). The goal of each episode is to keep the pole upright as long as possible, and the reward is $+1$ for each step taken. We set the number of episodes $K=800$. After episode $\Bar{K}=400$, the shift occurs by adding Gaussian noise $\mathcal{N}(0,0.15)$ to the cart and pole angular velocity features.

\textbf{State}: The state $s_t \in \mathbb{R}^4$ at time $t$ consists the following positions and velocities: (1) the cart position can take values between $(-4.8,4.8)$, but the episode terminates if the cart leaves the $(-2.4, 2.4)$ range; (2) the cart velocity can take values between $(-\infty,\infty)$; (3) the pole angle can be observed between $(-0.418,0.148)$ radians, but the episode terminates if the pole angle is not in the range $(-0.2095, 0.2095)$; (4) the pole angular velocity can take values between $(-\infty,\infty)$.

\textbf{Action}: The action $a_t\in \{0,1\}$ at time $t$ indicates the direction of the fixed force with which the cart is driven, where $0$ represents pushing the cart to the left and $1$ represents pushing the cart to the right.

\textbf{State Transition Function}: From the state $s_t$, the cart changed its positions and velocities to state $s_{t+1}$ by action $a_t$, with additional Gaussian noise $\epsilon \sim \mathcal{N}(0,0.15)$ to the cart and pole angular velocity features. All observations are assigned a uniformly random value in $(-0.05, 0.05)$. The episode ends if any one of the following occurs: (1) Termination: the pole angle is over $\pm 0.2095$ or the cart position is over $\pm 2.4$ (center of the cart reaches the edge of the display); (2) Truncation: episode length is greater than $200$.

\textbf{Reward Function}: Since the goal is to keep the pole upright as long as possible, a reward $r_t$ is added by $+1$ for step $t$, including the termination step. The reward threshold is $195$.

\textbf{Baselines}: We compare our method with: (1) Random policy; (2) Deep Q-Network (DQN)~\citep{mnih2013Atari,paszke2019torch}; (3) Deep Q-Network with UCB exploration (DQN-UCB). We adapt the hashing technique of~\citet{tang2017countdeep} to count the number of visited pairs $N_h(s,a)$ on this continuous state space. In particular, count-based exploration uses a static hashing to map continuous states into discrete states and then counts the number of times a given state has been visited. After that, the UCB algorithms are trained with a bonus reward considering the number of times we have visited the state. This bonus reward plays the role of exploration.

\subsubsection{COVID-19 patient hospital allocation}
\textbf{Notation}:
\begin{itemize}
    \item $K$: Number of hospitals.
    \item $T$: Number of days in the planning horizon.
    \item $c_i$: Capacity of hospital $i$, for $i=1,\cdots,K$.
    \item $N_t$: Number of arriving COVID-19 patients in the system on day $t$.
    \item $\{A_t\}_{t\in [T]}$: Stochastic process representing the number of arriving COVID-19 patients over time.
    \item $d_{i,t}$: Number of non-COVID patients in hospital $i$ on day $t$.
    \item $\{D_{i,t}\}_{t\in [T]}$: Stochastic process representing the number of non-COVID patients in hospital $i$ over time.
    \item $a_{i,t}$ (action): Number of arriving COVID-19 patients allocated to hospital $i$ on day $t$.
    \item $y_{i,t}$: COVID-19 patient occupancy in hospital $i$ on day $t$.
    \item $o_{i,t}$: Total occupancy in hospital $i$ on day $t$, defined as $o_{i,t} = d_{i,t} + y_{i,t}$.
    \item $L_i$: Random variable representing patient length of stay.
    \item $p(L_i\geq t)$: Probability that a patient at hospital $i$ stays at least $t$ days.
    \item $\beta_i$: Parameter for updating the estimated COVID-19 occupancy.
    \item $s_t$: State of the system at time $t$.
\end{itemize}
\textbf{State}: State $s_t\in \mathbb{R}^{2K+1}$ at time $t$ consists of the COVID-19 and non-COVID-19 occupancies in each hospital, along with the number of arriving patients:
\begin{align}
    s_t = (y_{1,t},y_{2,t}, \cdots, y_{K,t};\quad d_{1,t},d_{2,t}, \cdots, d_{K,t};\quad  N_t).
\end{align}

\textbf{Action}: Action $a_t \in \mathbb{R}^K$ at time t consists of the number of COVID-19 patients allocated to each hospital:
\begin{align}
    a_t = (a_{1,t}, a_{2,t}, \cdots, a_{K,t})
\end{align}
subject to the constraints: $\sum_{i=1}^N a_{i,t} = N_t$ and $a_{i,t} \geq 0, \forall i=1\cdots K$. To allocate $N_t$ patients to satisfy this constraint, we use an oracle function from $Q$-values, which receives $N_t$ patients and $Q$-values, then outputs action $a_t$ to $K$ hospitals~\citep {zuo2021combinatorialmultiarmedbanditsresource,li2024deepreinforcementlearningefficient}.

\textbf{State Transition Function}: 
The number of arriving COVID-19 patients is sampled from the corresponding stochastic process:
\begin{align}
    N_{t+1} \sim A_{t+1}.
\end{align}

The non-COVID-19 occupancy is sampled from the corresponding stochastic process (can be deterministic):
\begin{align}
    d_{i,t+1} \sim D_{i,t+1}.
\end{align}

The COVID-19 occupancy update:
\begin{align}
    y_{i,t+1} = \beta_i \cdot y_{i,t} + a_{i,t+1},
\end{align}
where $\beta_i \in \mathbb{R}^K$ is provided by the environment, which can be done by minimizing the difference between the estimated and actual COVID-19 occupancies from the real-world dataset: COVID-19 Reported Patient Impact and Hospital Capacity by Facility, provided by the U.S. Department of Health \& Human Services over $T=1274$ days from 2020 to 2024. We use the $K=40$ dataset collected from Texas hospitals.

\textbf{Reward Function}: The number of overflows across all hospitals:
\begin{align}
    r_t = \sum_{i=1}^K -\max(0,o_{i,t+1}-c_i),
\end{align}
where $o_{i,t+1}=y_{i,t+1}+d_{i,t+1}$ is the total occupancy of hospital $i$ at time $t+1$. This represents the cost of a capacity shortage.

\textbf{Baselines}: We compare our method with: (1-2) combinatorial resource allocation UCB methods, i.e., UCB\_RA and CUCB\_RA~\citep{zuo2021combinatorialmultiarmedbanditsresource}. And our Deep Q-learning extension, including (3) Deep Q-learning only (CNeural\_RA, i.e., DQN)~\citep{mnih2013Atari,paszke2019torch}, and (4) Deep Q-learning with UCB exploration (Q-learning, i.e., DQN-UCB).

\newpage
\subsubsection{Source code and computing systems}
Our source code includes the dataset scripts, setup for the environment, and our provided code (details in README.md). We run our code on a single GPU: NVIDIA RTX~A6000-49140MiB with 8-CPUs: AMD Ryzen Threadripper 3960X 24-Core with 8GB RAM per each and require 10GB available disk space for storage. 

\subsection{Demo notebook code for Algorithm~\ref{alg:episodic}}\label{apd:code}
\begin{tabular}{cc}
\begin{minipage}{0.1\linewidth}
\end{minipage}&
\begin{minipage}{0.96\linewidth}
\begin{minted}
[
frame=lines,
framesep=2mm,
baselinestretch=1,
bgcolor=LightGray,
fontsize=\fontsize{8.2pt}{8.2pt},
linenos
]
{python}
from rlberry.agents import IncrementalAgent, DiscreteCounter

class Ours(IncrementalAgent):
    def __init__(self):
        H = self.horizon
        S = self.env.observation_space.n
        A = self.env.action_space.n
        self.nu_kde = KernelDensity()
        self.de_kde = KernelDensity()
        self.list_nu_samples, self.list_de_samples = [], []
        # (s, a) visit counter
        self.N_sa = np.zeros((H, S, A))
        self.counter = DiscreteCounter(H, A)
        # Value functions
        self.V = np.ones((H+1, S))
        self.V[H, :] = 0
        self.Q = np.ones((H, S, A))
        self.Q_bar = np.ones((H, S, A))
        for hh in range(self.horizon):
            self.V[hh, :] *= (self.horizon-hh)
            self.Q[hh, :, :] *= (self.horizon-hh)
            self.Q_bar[hh, :, :] *= (self.horizon-hh)
        r_range = self.env.reward_range[1] - self.env.reward_range[0]
        self.v_max = np.zeros(self.horizon)
        for hh in reversed(range(self.horizon-1)):
            self.v_max[hh] = r_range + self.gamma*self.v_max[hh+1]

    def _get_action(self, state, hh=0):
        return self.Q_bar[hh, state, :].argmax()

    def _compute_bonus(self, n, hh, likelihood):
        bonus = self.bonus_scale_factor * np.sqrt(1.0 / n) + self.v_max[hh] / n
        bonus = min(bonus, self.v_max[hh])
        return bonus/likelihood

    def _update(self, state, action, next_state, reward, hh, likelihood):
        self.N_sa[hh, state, action] += 1
        nn = self.N_sa[hh, state, action]
        alpha = (self.horizon+1.0)/(self.horizon + nn)
        bonus = self._compute_bonus(nn, hh, likelihood)
        target = reward + bonus + self.gamma*self.V[hh+1, next_state]
        self.Q[hh, state, action] = (1-alpha)*self.Q[hh, state, action] + alpha * target
        self.V[hh, state] = min(self.v_max[hh], self.Q[hh, state, :].max())
        self.Q_bar[hh, state, action] = self.Q[hh, state, action]

    def _run_episode(self):
        # interact for H steps
        episode_rewards = 0
        state = self.env.reset()
        for hh in range(self.horizon):
            action = self._get_action(state, hh)
            next_state, reward, done, _ = self.env.step(action)
            episode_rewards += reward
            value1 = np.array([next_state, state, action])
            value2 = np.array([state, action])
            density_ratio = np.exp(self.nu_kde.score_samples(value1))
                /(np.exp(self.de_kde.score_samples(value2))
            self.counter.update(state, action)
            self._update(state, action, next_state, reward, hh, density_ratio)
            self.list_nu_samples.append(value1)
            self.list_de_samples.append(value2)
            self.nu_kde.fit(self.list_nu_samples[len(self.list_nu_samples)-100:])
            self.de_kde.fit(self.list_de_samples[len(self.list_nu_samples)-100:])
            state = next_state
            if done:
                break
        return episode_rewards
\end{minted}
\end{minipage}
\end{tabular}

\subsection{Additional results}\label{apd:add_results}
\subsubsection{Evaluation across different shift intensities}
\begin{figure}[ht!]
    \centering
    \includegraphics[width=1.0\linewidth]{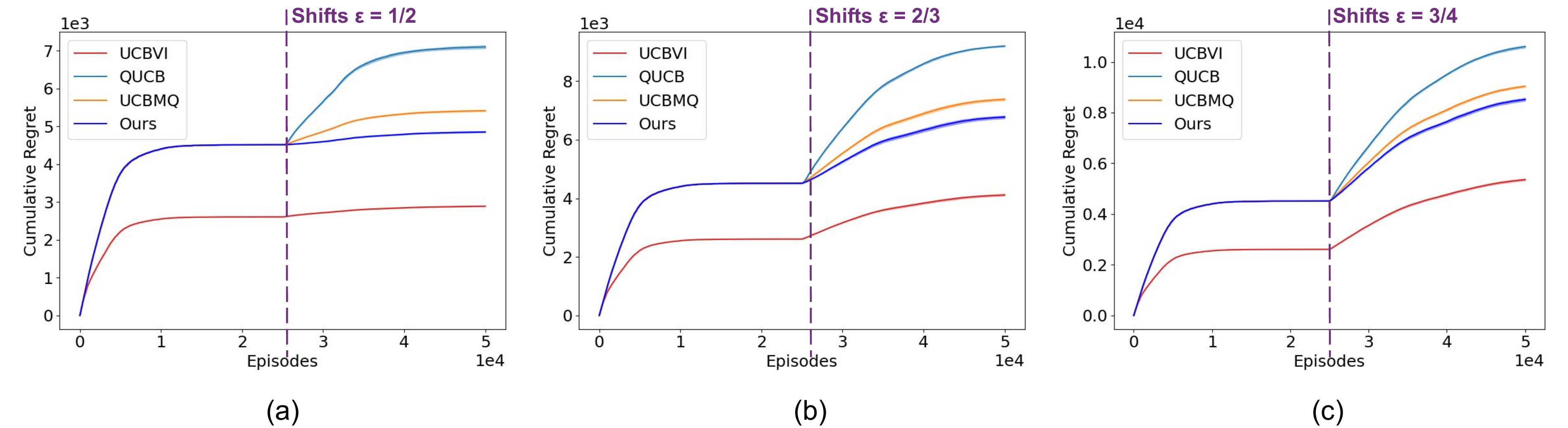}
    \caption{Cumulative regret comparison on Frozen-Lake with $K=50000$ average over $10$ runs, transition noise $\epsilon=0$ before the shift $\bar{K}=25000$. After that, $\epsilon=\{1/2, 2/3, 3/4\}$ in Figures~(a), (b), and (c), respectively.}
    \label{fig:grid_world_diff_shift}
\end{figure}

To evaluate the robustness of our method, we additionally compare it with other baselines across different shift intensities. Specifically, we deploy all models and test their performance with different levels of transition noise $\epsilon$ in the Frozen-Lake task. Figure~\ref{fig:grid_world_diff_shift} shows our result with $\epsilon=0$ before the shift and $\epsilon=\{1/2, 2/3, 3/4\}$ after the shift in Figures~\ref{fig:grid_world_diff_shift}~(a), (b), and (c), respectively. Firstly, we can see that our method consistently outperforms QUCB and UCBMQ by having a lower cumulative regret across different shift intensities. This once confirms our theoretical and empirical results in the main paper. Secondly, since the agent will move in the direction by following the action with probability $1-\epsilon$ and move perpendicular to the intended direction with probability $\epsilon$, we can see that when the transition noise $\epsilon$ increases, the performance of all methods will be degraded accordingly. Third, we observe that when the shift is too severe, such as $\epsilon=3/4$ (i.e., moving randomly with probability $75\%$), all models will find it hard to converge, even with model-based RL like UCVI in Figure~\ref{fig:grid_world_diff_shift}~(c).

\subsubsection{Evaluation across different types of environment shifts}
\begin{figure}[ht!]
    \centering
    \includegraphics[width=1.0\linewidth]{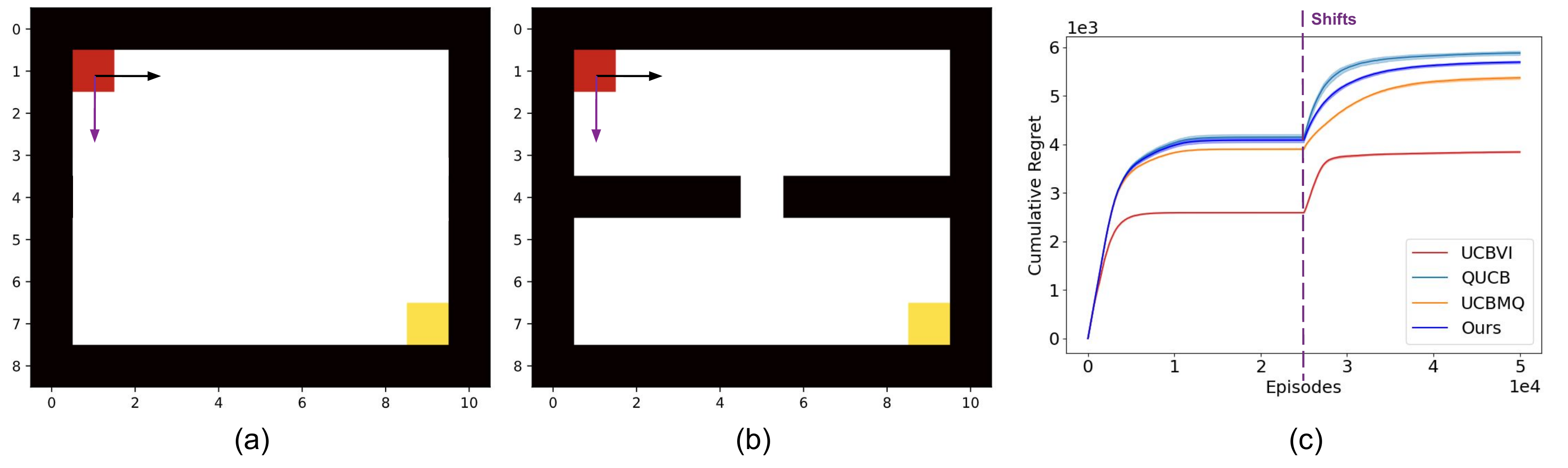}
    \caption{(a) The one-room GridWorld task: the starting state is shown in red, the rewarding state is shown in yellow, and the transitions are noisy with $\epsilon=0.2$; (b) The environment shift of (a) by changing to two-rooms GridWorld map; (c) Cumulative regret comparison average over $10$ runs, the map is one-room and two-rooms before and after the shift at $\bar{K}=25000$, respectively.}
    \label{fig:grid_world_two_room}
\end{figure}

Next, we consider a more real-world shift setting by changing the map of the GridWorld task. In particular, the agent starts to interact with the environment in the one-room map in Figure~\ref{fig:grid_world_two_room}~(a) (similar to the environment before the shift in Figure~\ref{fig:grid_world_diff_shift}). After that, the shift occurs in the episode $\bar{K}=25000$, the environment changes to a two-rooms map to interact with the agent in Figure~\ref{fig:grid_world_two_room}~(b). Since the map suddenly changes, several optimal directions in one-room map from the starting point (red) to the ending point (yellow) will be blocked in two-rooms map. This leads to all methods suffering from low rewards (high regrets) between episodes $25000$ and $30000$ in Figure~\ref{fig:grid_world_two_room}~(c). After that, they can adapt to the new map and converge with an optimal policy. Generally, we observe that the performance across the methods is quite similar to the previous results. Notably, the adaptation ability of our method is still better than QUCB, confirming the effectiveness of our UCB exploration in the non-stationary RL under distribution shifts.

\subsubsection{Additional comparison with model-based RL}
\begin{figure}[ht!]
    \centering
    \includegraphics[width=1.0\linewidth]{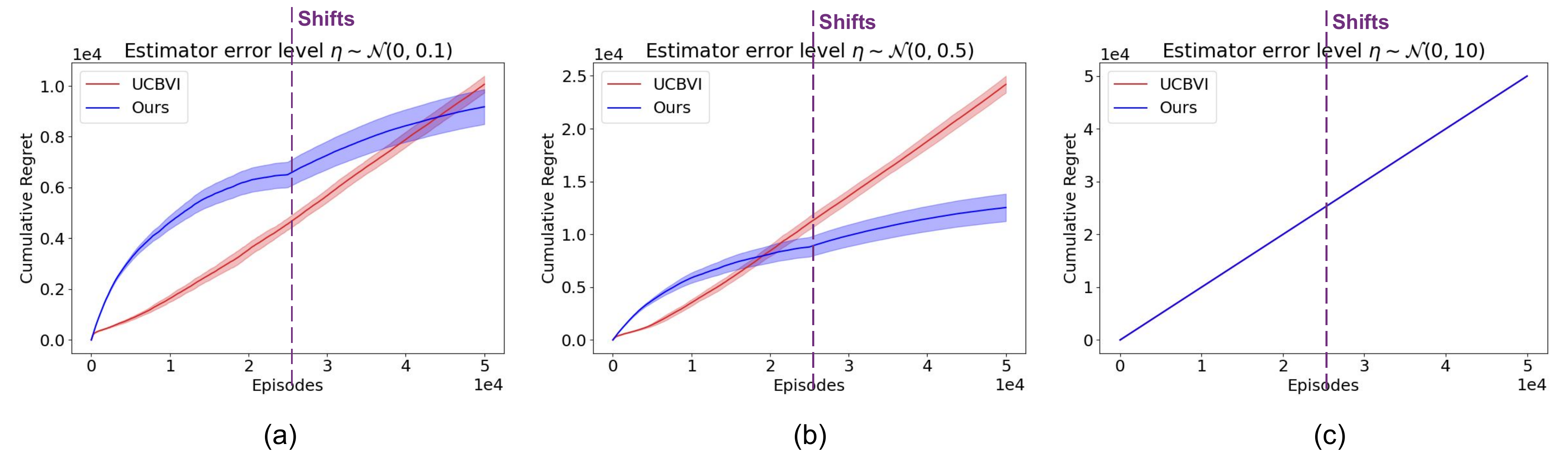}
    \caption{Cumulative regret comparison on Frozen-Lake with $K=50000$ average over $10$ runs, transition noise $\epsilon=0$ and $\epsilon=2/3$ before and after the shift at $\bar{K}=25000$. The transition estimator error level $\eta \sim  \mathcal{N}(0, std)$, where $std = \{0.1, 0.5, 10\}$ in Figures~(a), (b), and (c), respectively.}
    \label{fig:vsUCBVI}
\end{figure}
Finally, we provide further ablation studies for comparison with model-based RL. Recall that our method does not explicitly model the transition operator $\mathbb{P}_h$ like the model-based method (e.g., UCBVI, UCBMQ, etc.), which needs to store and iterate through all possible $(s',s,a) \in \mathcal{S} \times \mathcal{S} \times \mathcal{A}$ tuples. On the one hand, this helps our method avoid a high computational burden, as discussed in Remark~\ref{rmk:complexity} and empirically shown in Figure~\ref{fig:complexity}. On the other hand, this helps our method avoid heavy dependence on the transition estimator, which may be inaccurate and lead to poor regret performance in model-based RL. Indeed, we provide an analysis on how the transition estimator error affects regret results between our method versus the model-based UCBVI as follows.

We inherit the experimental setting on Frozen-Lake in Figure~\ref{fig:grid_world_diff_shift}~(b). In UCBVI, we denote $\hat{\mathbb{P}}_h$ as the transition estimator of the true underlying transition function $\mathbb{P}_h$. Hence, the transition estimator error is $||\hat{\mathbb{P}}_h-\mathbb{P}_h||$. To control the level of estimator error, we add a noise hypothesis $\eta$ to the transition estimator, i.e., $\hat{\mathbb{P}}_h + \eta$ for every episode $h\in [H]$, where the r.v. $\eta \sim \mathcal{N}(0, std)$. Then, we use this transition estimator to run with both UCBVI and our algorithm. Since the transition estimator is the same, we can guarantee that the two methods have the same transition estimation error level.

We summarize the results in Figure~\ref{fig:vsUCBVI}, where different sub-figures represent different levels of estimator error by controlling the level of $\eta$. Firstly, we observe that the higher the level of randomness in $\eta$, i.e., the higher the transition estimator error, leading to worse performance for both methods. Moreover, if this randomness is too high, then both methods result in a linear regret in Figure~\ref{fig:vsUCBVI}~(c). Secondly, UCBVI is more sensitive to the transition estimator error than our method, causing UCBVI to become worse than our method when the transition estimator error increases. For example, when $\eta \sim \mathcal{N}(0,0.1)$, UCBVI tends to be worse than us in the last round in Figure~\ref{fig:vsUCBVI}~(a). And when the error level of the estimator increases, that is, $\eta \sim \mathcal{N}(0,0.5)$, we can clearly observe that UCBVI is worse than our method with a much higher cumulative regret in Figure~\ref{fig:vsUCBVI}~(b). Therefore, we can conclude that besides the computational limitation, model-based RL, such as UCBVI, is more sensitive to the transition estimator than our method. This leads to when the transition estimator error is high in some environments, our method could potentially bring out a better performance in both computational and regret (i.e., reward) performance than model-based RL. 

\subsubsection{Density estimation ablation studies}
\begin{table}[ht!]
    \vspace{-0.2in}
    \centering
    \scalebox{0.8}{
    \begin{tabular}{ccccc}\\
    \toprule  
    Method & $t=T/3$ (shift) & $t=2T/3$ (shift) & $t=3T/4$ & $t=T$\\\midrule
    QUCB & 4512.0 $\pm$ 0.0 & 7571.0 $\pm$ 40.5 & 8129.0 $\pm$ 44.6 & 8951.0 $\pm$ 36.9\\
    UCBVI  & 4513.5 $\pm$ 0.0 & 5737.25 $\pm$ 8.8 & 6101.0 $\pm$ 13.9 & 6703.25 $\pm$ 13.8\\
    \midrule
    Ours (kernel)\\
    \midrule
    Linear & 4515.5 $\pm$ 0.0 & 5139.25 $\pm$ 9.6 & 5441.0 $\pm$ 13.0 & 5970.0 $\pm$ 13.0\\
    Cosine & 4515.5 $\pm$ 0.0 & 5137.0 $\pm$ 8.6 & 5442.5 $\pm$ 12.9 & 5974.5 $\pm$ 13.0\\
    Gaussian (default) & 4513.5 $\pm$ 0.0 & 5126.0 $\pm$ 9.6 & 5425.0 $\pm$ 13.0 & 5954.0 $\pm$ 13.7\\
    Exponential	& 4512.5 $\pm$ 0.0 & 5125.0 $\pm$ 8.8 & 5422.5 $\pm$ 12.5 & 5948.0 $\pm$ 13.5\\
    \midrule
    Ours (bandwidth)\\
    \midrule
    0.01 & 4513.5 $\pm$ 0.0	& 5598.0 $\pm$ 10.1 & 6055.25 $\pm$ 16.5 & 6501.5 $\pm$ 20.7\\
    0.1	& 4513.0 $\pm$ 0.0 & 5123.5 $\pm$ 8.8 & 5420.0 $\pm$ 12.6 & 5947.5 $\pm$ 13.7\\
    1.0 (default) & 4513.5 $\pm$ 0.0 & 5126.0 $\pm$ 9.6 & 5425.0 $\pm$ 13.0 & 5954.0 $\pm$ 13.7\\
    10 & 4513.5 $\pm$ 0.0 & 7526.0 $\pm$ 38.9 & 8121.25 $\pm$ 37.3 & 8949.0 $\pm$ 33.0\\
    \midrule
    Ours (window sizes)\\
    \midrule
    1 & 4513.0 $\pm$ 0.0 & 7514.25 $\pm$ 36.7 & 8122.0 $\pm$ 44.0 & 8943.0 $\pm$ 36.5\\
    10 & 4513.0 $\pm$ 0.0 & 6098.25 $\pm$ 10.1 & 6391.0 $\pm$ 16.6 & 7007.0 $\pm$ 21.4\\
    100 (default) & 4513.5 $\pm$ 0.0 & 5126.0 $\pm$ 9.6 & 5425.0 $\pm$ 13.0	& 5954.0 $\pm$ 13.7\\
    1000 & 4513.5 $\pm$ 0.0 & 5049.0 $\pm$ 9.0 & 5311.25 $\pm$ 12.5 & 5729.25 $\pm$ 12.8\\
    \bottomrule
    \end{tabular}}
    \caption{\small{Ablation studies in cumulative regret (lower is better) across different density function’s parameters, including kernel type, bandwidth, and window size. Results are reported over 10 runs. Experiments correspond to our Fig~\ref{fig:regret}~(b) on Frozen-Lake with two shifts at time $t=T/3$ and $t=2T/3$, where $T=60000$.}}
    \label{tab:density_ablation}
\end{table}

Table~\ref{tab:density_ablation} utilizes different kernel density function parameters, including various kernel types, different bandwidth sizes, and different window sizes, in the Frozen-Lake environment. We observe that our method is quite robust across different kernel types (e.g, Gaussian, Exponential, Cosine, and Linear) and the bandwidth between the range of (0.01, 1.0). If the bandwidth is too large (e.g., 10), then the KDE is oversmoothed and biased, leading to a degradation in regret performance and resulting in a similar result with QUCB. Similarly, if the window size is too small (e.g., (1,10)), the density function will also be biased by only having a small observation. That said, with a sufficient number of samples (100,1000), our method consistently outperforms QUCB and UCBMQ.

\subsubsection{Extension to continuous action space on MuJoCo’s locomotion}
\begin{table}[ht!]
    \centering
    \vspace{-0.2in}
    \scalebox{0.8}{
    \begin{tabular}{ccccc}\\
    \toprule  
    Method on Hopper & $t=T/2$ (shift) & $t=2T/3$ (shift) & $t=3T/4$ & $t=T$\\\midrule
    DQN	& 179.4 $\pm$ 1.0 & 221.0 $\pm$ 1.4 & 295.3 $\pm$ 1.5 & 410.2 $\pm$ 2.3\\
    DQN-UCB	& 211.0 $\pm$ 0.8 & 257.2 $\pm$ 1.2 & 331.1 $\pm$ 1.3 & 509.0 $\pm$ 1.7\\
    Ours & 211.8 $\pm$ 0.8 & 397.5 $\pm$ 1.2 & 516.4 $\pm$ 1.2 & 689.6 $\pm$ 1.4\\
    \midrule
    Method on Halfcheetah\\
    \midrule
    DQN	& 1,410.1 $\pm$ 2.4	& 1,595.6 $\pm$ 2.6 & 1,800.8 $\pm$ 3.0	& 1,967.2 $\pm$ 3.8\\
    DQN-UCB	& 1,526.1 $\pm$ 2.0	& 1,733.4 $\pm$ 2.3	& 1,920.7 $\pm$ 2.6	& 2,125.0 $\pm$ 2.5\\
    Ours & 1,526.7 $\pm$ 2.0 & 2,017.6 $\pm$ 2.2 & 2,293.6 $\pm$ 2.3 & 2,694.9 $\pm$ 2.2\\
    \midrule
    Method on Walker\\
    \midrule
    DQN	& 288.5 $\pm$ 0.9 & 319.1 $\pm$ 1.0	& 380.7 $\pm$ 1.2 & 477 $\pm$ 1.4\\
    DQN-UCB	& 323.3 $\pm$ 0.7 & 367.0 $\pm$ 0.7	& 433.3 $\pm$ 0.8 & 595 $\pm$ 0.8\\
    Ours & 323.8 $\pm$ 0.6 & 411.6 $\pm$ 0.6 & 631.6 $\pm$ 0.6 & 755 $\pm$ 0.7\\
    \bottomrule
    \end{tabular}}
    \caption{\small{Experimental results in average return (higher is better) on MuJoCo’s locomotion, including hopper, halfcheetah, and walker environments across $T=200000$. Results are reported over 10 runs.}}
    \label{tab:contonous_action}
\end{table}

Although we focus on Q-learning (which is designed for discrete action spaces), our framework can also be extended for continuous action spaces by using the Discretizing Continuous Action Space method~\citep{tang2020Discretizing}. Specifically, given action space $\mathcal{A}=[-1,1]^m$ in Mojuco Locomotion, we discretize each dimension of the action space into $K=100$ equally spaced atomic actions (bins). The set of atomic actions for any dimension $i$ is $\mathcal{A}=[\frac{2j}{K-1}-1]_{j=0}^{K-1}$. During inference time, we convert the discrete Deep-Q-Learning outputs to continuous w.r.t. the mean of each bin. Table~\ref{tab:contonous_action} shows that our method also outperforms other baselines, achieving a higher average return over 20 evaluation episodes across MuJoCo Locomotion’s tasks. This result is consistent with the lowest cumulative regret, confirming the effectiveness of our method’s shift awareness under distribution shifts.
\end{document}